\documentclass[conference]{IEEEtran}
\IEEEoverridecommandlockouts

\usepackage{cite}
\usepackage{amsmath,amssymb,amsfonts}
\usepackage{algorithm}
\usepackage{algorithmic}
\usepackage{graphicx}
\usepackage{textcomp}
\usepackage{xcolor}
\usepackage{booktabs}
\usepackage{multirow}
\usepackage[table]{xcolor}
\usepackage{makecell}
\usepackage{adjustbox}
\usepackage{tikz}
\usetikzlibrary{arrows.meta,backgrounds,fit,positioning,decorations.pathmorphing}
\usepackage{mathtools}
\usepackage{amsthm}
\usepackage{bm}
\usepackage{xspace}
\usepackage{marvosym}
\usepackage{hyperref}
\usepackage{cleveref}
\usepackage{balance}

\def\BibTeX{{\rm B\kern-.05em{\sc i\kern-.025em b}\kern-.08em
    T\kern-.1667em\lower.7ex\hbox{E}\kern-.125emX}}

\definecolor{algblue}{RGB}{36,99,175}
\definecolor{algred}{RGB}{181,79,57}
\definecolor{alggreen}{RGB}{39,130,91}

\newtheorem{theorem}{Theorem}[section]

\theoremstyle{definition}
\newtheorem{definition}[theorem]{Definition}
\theoremstyle{remark}

\newcommand{\std}[1]{{\tiny$\!\pm\!$#1}}
\newcommand{\rank}[2]{%
  \ifnum#1=1{\textbf{\textcolor{red!70}{#2}}}%
  \else\ifnum#1=2{\textcolor{blue!70}{\underline{#2}}}%
  \else #2%
  \fi\fi%
}

\newcommand{\Method}{\textsc{AOSNet}\xspace}
\newcommand{\PaperTitle}{Adaptive Oscillatory-State Alignment for Time Series Forecasting}
\newcommand{\best}[1]{\cellcolor{gray!20}{#1}}
\newcommand{\bx}{\mathbf{x}}
\newcommand{\bX}{\mathbf{X}}
\newcommand{\by}{\mathbf{y}}
\newcommand{\bY}{\mathbf{Y}}
\newcommand{\bP}{\mathbf{P}}
\newcommand{\R}{\mathbb{R}}
\newcommand{\AlgComment}[1]{\hfill{\footnotesize$\triangleright$~\textit{#1}}}
\renewcommand{\IEEEauthorrefmark}[1]{\textsuperscript{#1}}

\begin{document}

\title{\PaperTitle\thanks{}}


  \author{
  \IEEEauthorblockN{
  Zhangyao Song\IEEEauthorrefmark{1},
  Chaofeng Qu\IEEEauthorrefmark{1},
  Chao Zha\IEEEauthorrefmark{2},
  Xiaoyu Zhao\IEEEauthorrefmark{1},
  Yinfei Xu\IEEEauthorrefmark{3},
  Tao Guo\IEEEauthorrefmark{1}\IEEEauthorrefmark{\Letter}
  }
  \IEEEauthorblockA{\IEEEauthorrefmark{1}School of Cyber Science and Engineering, Southeast University, Nanjing, China}
  \IEEEauthorblockA{\IEEEauthorrefmark{2}State Key Laboratory of Blockchain and Data Security, Zhejiang University, Hangzhou, China}
  \IEEEauthorblockA{\IEEEauthorrefmark{3}School of Information Science and Engineering, Southeast University, Nanjing, China}
  }
\maketitle

\begin{abstract}
  Long-term time series forecasting benefits from inductive biases that expose recurring temporal structure. Existing periodic forecasting methods typically model recurrence through predefined periods, global spectral components, or fixed learnable templates. However, real-world temporal dynamics are rarely rigidly periodic: around a nominal cycle, oscillatory behavior often exhibits \emph{non-rigid periodicity} (NRP), where cycle magnitude, cycle alignment, and local cycle duration vary over time. Under these conditions, fixed-template periodic modeling can become fundamentally mismatched to the underlying temporal states. We propose \Method, a Hilbert-guided forecasting framework that reformulates periodic forecasting from fixed template matching to adaptive oscillatory-state alignment. \Method extracts analytic-signal descriptors from both the observed sequence and a learnable global oscillatory prior, then adaptively aligns local states through a descriptor-conditioned gate that selectively preserves reliable observations while softly correcting mismatched regions. The learned prior serves not as a rigid repeated template but as a flexible oscillatory reference interpreted through local state dynamics. Experiments on eight public benchmarks and two cloud workload traces demonstrate leading or highly competitive accuracy with a compact model size and low inference latency, supporting repeated forecasting settings such as capacity planning and autoscaling. Controlled synthetic studies that isolate cycle-magnitude and cycle-alignment variation and combine them with cycle-duration changes show that the advantage of oscillatory-state alignment increases as NRP intensifies.
\end{abstract}


\section{Introduction}
\label{sec:introduction}

Time series forecasting is a fundamental problem in machine learning with broad applications in energy systems, transportation, finance, climate science, and cloud resource management~\cite{Wen-2022-Transformerstimeseries,Qiu-2024-TFB,Shao-2025-ExploringProgressMultivariate,Guo-2026-SurveyDeepLearning_TSF,Zhou-2021-Informerefficienttransformer,Zeng-2023-DLinear,Zhang-2025-MultiPeriodLearning,Chen-2023-FuXi}. In long-term forecasting, predictive performance depends not only on modeling short-term dependencies but also on capturing recurring temporal structure over extended horizons~\cite{Wu-2021-Autoformer,Nie-2023-PacthTST,Liu-2024-iTransformer}. This is especially important for operational workload forecasting, where inaccurate long-horizon predictions can affect capacity planning, autoscaling, and resource allocation. Beyond accuracy, such operational settings impose efficiency constraints because forecasting models may be invoked repeatedly in capacity-planning and autoscaling loops. Consequently, recent research increasingly relies on explicit temporal inductive biases, including decomposition methods, spectral representations, and learnable periodic priors~\cite{Ye-2024-FAN,Liu-2025-TimeBridge,Song-2026-DSTND}.

Despite their architectural differences, most existing periodic forecasting methods share a common underlying assumption: recurrence can be represented as a fixed repeated template. Some approaches explicitly estimate a dominant period and align observations according to cycle indices~\cite{Lin-2024-CycleNet,Lin-2025-TQNet}; others model periodicity through global frequency components~\cite{Zhou-2022-FILM,Ye-2024-FAN,Wang-2025-FreDF,Fei-2025-Amplifier} or learnable periodic embeddings. Decomposition methods separate seasonal and trend components under similar stationarity assumptions~\cite{Zeng-2023-DLinear,Wang-2024-TimeMixer,Yu-2024-Leddam,Deng-2024-ParsimonyCapabilityDecomposition}. These methods have demonstrated strong empirical performance, suggesting that explicit periodic structure is indeed a useful forecasting prior.

The difficulty, however, is not merely that the correct period may be unknown. Many real-world series contain recurring behavior around a nominal period, but the cycles themselves are non-rigid. Electricity demand may retain a daily rhythm while cycle intensity varies with weather and human activity; traffic flows may exhibit delayed or advanced cycle peaks under holidays; cloud workloads may preserve diurnal usage patterns while shifting under scaling events or service bursts; and environmental signals may contain cycles that locally stretch or compress. We formalize these cases as one common problem: \emph{non-rigid periodicity} (NRP).

\begin{definition}[Non-Rigid Periodicity]
  Given a nominal period $P_0$ and reference phase $\theta_0(t)=2\pi t/P_0$, a signal exhibits \emph{non-rigid periodicity} if its local cycle state
  \[
    \boldsymbol{\xi}(t)=\big(A(t),\,\delta(t),\,P(t)\big)
  \]
  varies over time, where $A(t)$ is the instantaneous envelope, $\delta(t)=\phi(t)-\theta_0(t)-\phi_0$ is the phase residual relative to the nominal cycle, $\phi_0$ is a constant phase offset, and $P(t)=2\pi/\omega(t)$ with $\omega(t)=\mathrm{d}\phi/\mathrm{d}t$ is the local cycle duration. This unified view yields three subproblems:
  \emph{cycle-magnitude variation} ($\mathrm{d}A/\mathrm{d}t\neq0$), where cycles repeat with changing strength;
  \emph{cycle-alignment variation} ($\delta(t)$ is non-constant), where peaks and troughs arrive earlier or later than the nominal cycle;
  and \emph{cycle-duration variation} ($\mathrm{d}P/\mathrm{d}t\neq0$), where consecutive cycles are locally stretched or compressed.
\end{definition}

These subproblems are coupled in practice---for instance, a time-varying local cycle duration necessarily induces cycle-alignment drift---but they describe distinct ways in which a period-indexed template can fail. Fixed-template or fixed-period methods implicitly assume that the cycle state $\boldsymbol{\xi}(t)$ is stable: the same nominal phase should retrieve the same amplitude, alignment, and cycle duration. This assumption breaks under NRP, motivating the adaptive oscillatory-state alignment of \Method.

\begin{figure}[t]
  \centering
  \includegraphics[width=\columnwidth]{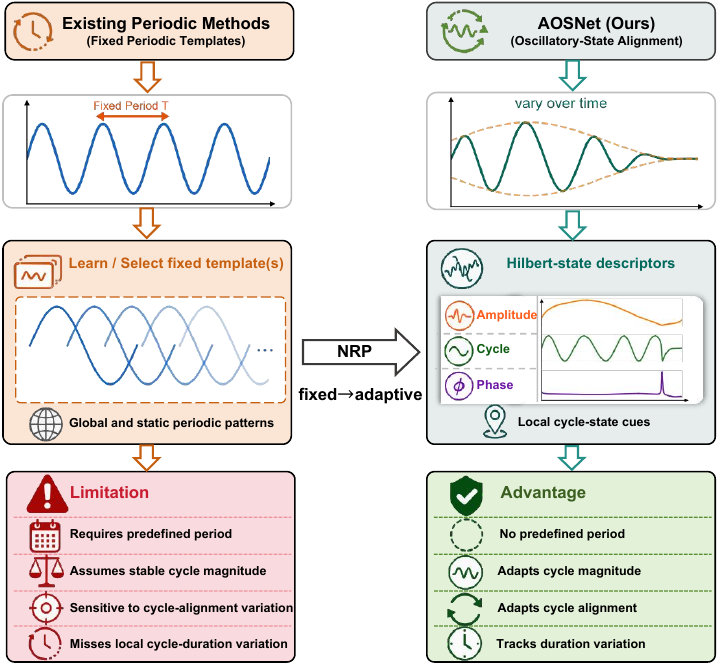}
  \caption{Motivation of \Method. Existing periodic methods typically learn or select fixed periodic templates, which can be sensitive to NRP: cycle magnitude, alignment, and duration may all vary over time. \Method instead represents temporal structure as adaptive local oscillatory states derived from amplitude, phase, and instantaneous-frequency descriptors.}
  \label{fig:motivation}
\end{figure}

This observation suggests a different view of periodic forecasting. Instead of treating periodicity as repeated template retrieval, forecasting should operate in an oscillatory-state space, where recurrence is represented by evolving local dynamical states. The key object is not a globally repeated cycle, but a local oscillatory state describing how strongly the signal oscillates, where it currently lies within the oscillation, and how rapidly the oscillation evolves. These quantities correspond to the amplitude envelope, instantaneous phase, and instantaneous frequency in analytic signal analysis~\cite{Gabor-1946-TheoryCommunication,Boashash-1992-EstimatingInterpreting}. The Hilbert transform provides differentiable descriptors of these states directly from real-valued sequences, without requiring period selection through prior knowledge, autocorrelation search, or frequency truncation. Fig.~\ref{fig:motivation} illustrates this motivation.

Based on this perspective, we propose \Method, a Hilbert-guided forecasting framework for adaptive oscillatory-state alignment. Instead of constructing a cycle table indexed by a predefined period, \Method learns a channel-wise global oscillatory prior shared across samples and constructs local oscillatory-state descriptors from both the observed sequence and the prior. A descriptor-conditioned adaptive gate then performs local state alignment by selectively preserving reliable observations while softly correcting mismatched regions toward the learned reference. The learned prior is not used as a rigid repeated template; rather, it serves as a flexible oscillatory reference that is adaptively interpreted through local state descriptors.

Concretely, \Method derives log-amplitude, phase (sine and cosine), and instantaneous frequency from both the input and the global prior. A lightweight convolutional gate consumes these descriptors and produces a time- and channel-dependent fusion coefficient. After this Hilbert-guided correction, a compact dual-path backbone combines an attention path for cross-variate refinement with a base path for stable temporal projection. Overall, \Method is designed for a practical accuracy--efficiency trade-off: its Hilbert-domain alignment adds limited overhead, and the dual-path backbone maintains forecasting capacity with low inference latency.

Our contributions are summarized as follows.
\begin{itemize}
  \item We reformulate periodic forecasting from fixed template matching to adaptive oscillatory-state alignment, identifying NRP---cycle-magnitude, cycle-alignment, and cycle-duration variation---as the key mismatch source that fixed-template methods cannot accommodate.
  \item We propose \Method, which compares observations with a learnable global oscillatory prior through Hilbert-domain envelope, phase, and instantaneous-frequency descriptors, then adaptively aligns local states via a descriptor-conditioned gate.
  \item We demonstrate leading or highly competitive accuracy on eight public benchmarks and two cloud workload traces, while retaining practical efficiency in parameter count and inference latency. Controlled synthetic experiments isolate cycle-magnitude and cycle-alignment variation and evaluate cycle-duration changes in the combined setting, where \Method consistently outperforms fixed-template baselines with a growing advantage as NRP intensifies.
\end{itemize}

\section{Related Work}
\label{sec:related_work}

In this section, we review three lines of work most relevant to \Method: periodic and decomposition-based forecasting, frequency-domain and signal-processing methods, and multivariate dependency modeling. This discussion positions \Method against existing temporal-structure priors and clarifies how adaptive oscillatory-state alignment differs from fixed-template or global spectral modeling.

\subsubsection{Periodic and Decomposition-Based Forecasting}
Long-term forecasting has increasingly relied on explicit temporal-structure modeling. Attention- and convolution-based architectures enlarge the receptive field for long-range dependencies~\cite{Zhou-2021-Informerefficienttransformer,Wu-2021-Autoformer,Zhou-2022-FEDFormer,Wang-2023-MICN,Nie-2023-PacthTST}, while lightweight decomposition methods prove highly competitive~\cite{Zeng-2023-DLinear,Wang-2024-TimeMixer,Lin-2024-SparseTSF,Yu-2024-Leddam}. More recently, learnable periodic priors have emerged as strong forecasting inductive biases: CycleNet learns recurrent cycle templates~\cite{Lin-2024-CycleNet}, TQNet injects periodically shifted learnable vectors as temporal queries~\cite{Lin-2025-TQNet}, and MoFo models periodic patterns for long-term prediction~\cite{Ma-2025-MoFo}. These methods share a common assumption---recurrence is represented as a fixed repeated template indexed by a predefined or estimated period. When the underlying oscillatory dynamics exhibit NRP, i.e., cycle magnitude, alignment, or cycle duration varies over time, this fixed-template assumption becomes a fundamental limitation. \Method departs from this paradigm by performing adaptive oscillatory-state alignment through Hilbert-domain descriptors, preserving the benefit of explicit temporal priors without requiring globally stable periods.

\subsubsection{Frequency-Domain and Signal-Processing Methods}
Frequency-domain approaches exploit the spectral separability of periodic structure. FEDformer and FILM model long-range dependencies via frequency representations~\cite{Zhou-2022-FEDFormer,Zhou-2022-FILM}; FITS performs lightweight frequency interpolation~\cite{Xu-2024-FITS}; FAN addresses non-stationarity through frequency-adaptive normalization~\cite{Ye-2024-FAN}; FreDF learns directly in the frequency domain~\cite{Wang-2025-FreDF}; and Amplifier recovers neglected low-energy components~\cite{Fei-2025-Amplifier}. While effective, these methods operate on global or window-level spectral components and thus implicitly assume locally stationary frequency content. \Method instead derives time-local analytic-signal descriptors---envelope, instantaneous phase, and instantaneous frequency---via the Hilbert transform~\cite{Gabor-1946-TheoryCommunication,Boashash-1992-EstimatingInterpreting}. This enables modeling of cycle-state variation at each time step, providing finer-grained oscillatory-state information than global spectral decomposition.

\subsubsection{Multivariate Dependency Modeling}
Cross-variate dependency modeling complements temporal modeling in multivariate forecasting. Early Transformer variants mix temporal and channel dimensions through attention~\cite{Vaswani-2017-transformer,Zhou-2021-Informerefficienttransformer,Wu-2021-Autoformer,Song-2026-CTPNet}; PatchTST advocates channel independence for robustness~\cite{Nie-2023-PacthTST}; and subsequent work revisits the independence-versus-interaction trade-off~\cite{Han-2024-capacityrobustnesstrade}. Inverted or channel-token architectures---iTransformer~\cite{Liu-2024-iTransformer}, Crossformer~\cite{Zhang-2023-Crossformer}, HDMixer~\cite{Huang-2024-HDMixer}, SAMformer~\cite{Ilbert-2024-SAMformer}---further advance multivariate interaction learning. These methods focus on how variables interact after representation, yet the temporal signal fed into cross-variate modules may still carry non-rigid oscillatory distortions. \Method addresses this by first performing per-channel oscillatory-state alignment, so that downstream channel-token attention operates on structurally refined signals rather than raw observations.

\section{Preliminaries}
\label{sec:preliminaries}

This section briefly reviews the analytic-signal descriptors used by \Method and their FFT-based computation.

\subsection{Analytic Signal and Hilbert Descriptors}

For a real-valued signal $x(t)\in L^2(\R)$, the Hilbert transform is defined as the Cauchy principal value integral~\cite{Gabor-1946-TheoryCommunication}:
\begin{equation}
  \mathcal{H}\{x\}(t)
  =
  \frac{1}{\pi}\,\mathrm{p.v.}\!\int_{-\infty}^{\infty}\frac{x(\tau)}{t-\tau}\,\mathrm{d}\tau
  =
  \left(\frac{1}{\pi t}\right)*x(t),
  \label{eq:hilbert_continuous}
\end{equation}
where $*$ denotes convolution. In the frequency domain, the Hilbert transform has transfer function
\begin{equation}
  H(f)=\mathcal{F}\!\left\{\frac{1}{\pi t}\right\}=-i\,\operatorname{sgn}(f)=
  \begin{cases}
    -i, & f>0, \\
    0,  & f=0, \\
    +i, & f<0.
  \end{cases}
  \label{eq:hilbert_transfer}
\end{equation}
The analytic signal combines the original signal and its Hilbert transform:
\begin{equation}
  z(t) = x(t) + i\,\mathcal{H}\{x\}(t).
  \label{eq:analytic_def}
\end{equation}
Its spectrum is one-sided:
\begin{equation}
  Z(f)
  =
  \begin{cases}
    2X(f), & f>0, \\
    X(0),  & f=0, \\
    0,     & f<0,
  \end{cases}
  \label{eq:analytic_spectrum}
\end{equation}
where $X(f)=\mathcal{F}\{x\}(f)$. This representation provides local amplitude, phase, and frequency descriptors for oscillatory dynamics.

For a discrete real-valued sequence $\bx\in\R^L$, let $\mathbf{z}\in\mathbb{C}^L$ be its analytic signal. We derive three descriptors:
\begin{align}
  \mathbf{a} & = \log\left(|\mathbf{z}|+\epsilon\right),                 \\
  \mathbf{c} & = \cos(\angle\mathbf{z}), \qquad
  \mathbf{s}_{\phi} = \sin(\angle\mathbf{z}),                             \\
  \boldsymbol{\omega}_{\ell}
             & = \angle\!\left(z_{\ell}z_{\ell-1}^{*}\right),
  \quad \ell=1,\ldots,L-1.
  \label{eq:hilbert_descriptors}
\end{align}
The log-amplitude descriptor $\mathbf{a}$ captures the oscillation envelope, $(\mathbf{c},\mathbf{s}_{\phi})$ gives a bounded phase representation without explicit unwrapping, and $\boldsymbol{\omega}$ is a phase-increment descriptor of local oscillation speed. In \Method, these quantities are used as differentiable structural descriptors rather than strict physical instantaneous-frequency estimates. Real-world multivariate series are not necessarily narrow-band or mono-component, and finite-window FFT descriptors can contain boundary effects; the learned gate therefore decides how strongly these descriptors should influence alignment.

\begin{figure*}[t]
  \centering
  \includegraphics[width=0.8\textwidth]{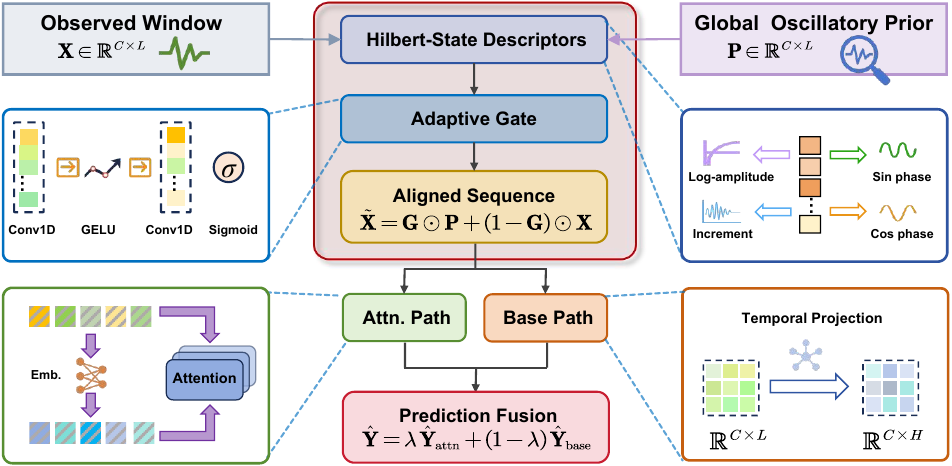}
  \caption{Overall architecture of \Method. Given an observed window, \Method introduces a learnable global oscillatory prior as a flexible reference. The AOS module extracts Hilbert-state descriptors from both the observation and the prior, estimates a descriptor-conditioned gate, and performs local oscillatory-state alignment. The aligned representation is then processed by a base path for stable per-channel extrapolation and an attention path for cross-variate refinement, whose predictions are fused by a learnable coefficient.}
  \label{fig:framework}
\end{figure*}

\subsection{FFT-Based Computation}

Algorithm~\ref{alg:analytic_signal} summarizes the analytic-signal computation for a single real-valued sequence. The frequency mask preserves the DC component, preserves the Nyquist component for even-length sequences, doubles positive frequencies, and removes negative frequencies.

\begin{algorithm}[t]
  \caption{FFT-Based Analytic Signal Computation}
  \label{alg:analytic_signal}
  \begin{algorithmic}[1]
    \REQUIRE Real-valued sequence $\bx=(x_0,\ldots,x_{L-1})\in\R^L$
    \ENSURE Analytic signal $\mathbf{z}\in\mathbb{C}^{L}$
    \STATE $\widehat{\bx}\leftarrow \mathcal{F}(\bx)$ \AlgComment{Fourier spectrum}
    \STATE $\mathbf{m}\leftarrow \mathbf{0}\in\R^L$ \AlgComment{Frequency mask}
    \IF{$L$ is even}
      \STATE $m_0\leftarrow 1$, $m_{L/2}\leftarrow 1$ \AlgComment{DC and Nyquist}
      \FOR{$k=1,\ldots,L/2-1$}
        \STATE $m_k\leftarrow 2$ \AlgComment{Positive frequency}
      \ENDFOR
    \ELSE
      \STATE $m_0\leftarrow 1$ \AlgComment{DC component}
      \FOR{$k=1,\ldots,(L-1)/2$}
        \STATE $m_k\leftarrow 2$ \AlgComment{Positive frequency}
      \ENDFOR
    \ENDIF
    \STATE $\mathbf{z}\leftarrow \mathcal{F}^{-1}(\mathbf{m}\odot\widehat{\bx})$ \AlgComment{One-sided inverse FFT}
    \RETURN $\mathbf{z}$
  \end{algorithmic}
\end{algorithm}

For descriptor extraction, amplitudes are clamped by $\epsilon$ before the logarithm, and the first phase-increment value is padded by replication. For a $C$-variate window, the FFT-based descriptor extraction costs $O(CL\log L)$ and introduces no recurrent or attention-style temporal dependency over the look-back window.

\section{Methodology}
\label{sec:methodology}

Given a multivariate historical window $\bX \in \R^{C \times L}$ with $C$ variables and look-back length $L$, point forecasting aims to predict the future horizon $\bY \in \R^{C \times H}$:
\begin{equation}
  f_{\theta}: \bX_{t-L+1:t} \mapsto \widehat{\bY}_{t+1:t+H},
\end{equation}
where $H$ is the forecasting length. This section presents \Method, a Hilbert-guided forecasting framework that reformulates periodic modeling from fixed template matching to adaptive oscillatory-state alignment. For notation clarity, we omit the time index and write the input as $\bX$ and the prediction as $\widehat{\bY}$. \Method follows the channel-token convention commonly used in multivariate forecasting: each variate is first represented as an individual temporal token, and cross-variate dependencies are refined after oscillatory-state alignment. Unlike fixed-template methods that retrieve cycle states by a predefined period index, \Method represents recurrence through adaptive oscillatory-state alignment guided by Hilbert-domain descriptors.

\subsection{Overview}

Before forecasting, each input window is temporally instance-normalized to reduce train-test statistical shifts, and the final prediction is inversely transformed to the original scale~\cite{Kim-2021-RevIN,Wu-2025-SRSNet}. The overall framework then contains three components, as illustrated in Fig.~\ref{fig:framework}. First, \Method maintains a learnable global oscillatory prior that provides a flexible oscillatory reference for the look-back window. Second, the adaptive oscillatory-state (AOS) module compares the normalized observation with this prior in a local oscillatory-state space, where cycle-magnitude, cycle-alignment, and cycle-duration changes are represented by Hilbert-state descriptors, and performs descriptor-conditioned alignment. Third, the aligned sequence is passed to a dual-path forecasting head that combines stable per-channel extrapolation with cross-variate refinement. This design keeps the main temporal inductive bias in the alignment module while avoiding hard period indexing.

\subsection{Global Oscillatory Prior}

\Method introduces a learnable channel-wise prior $\bP \in \R^{C \times L}$, shared across all samples and optimized jointly with the forecasting model. The prior has the same temporal shape as the look-back window, so it can serve as an alignment input for each channel. Unlike fixed periodic templates indexed by a predefined period, $\bP$ is not constrained to repeat with any cycle length. It acts as a flexible oscillatory reference whose contribution is interpreted locally by the AOS module. This distinction is important under NRP: when cycle magnitude, alignment, or duration changes over time, a useful reference should be modulated by local state evidence rather than imposed as a rigid template.

\subsection{Adaptive Oscillatory-State Alignment}

Given the observation $\bX$ and the prior $\bP$, the AOS module performs three operations: it extracts Hilbert-state descriptors, estimates a descriptor-conditioned gate, and aligns the observation to the prior through local interpolation.

\subsubsection{Hilbert-State Descriptor Extraction}

Using the analytic-signal descriptor operator defined in Sec.~\ref{sec:preliminaries}, we compute Hilbert-state descriptors for both the observed sequence and the global oscillatory prior:
\begin{equation}
  \mathcal{D}(\bX)
  =
  \{\mathbf{a}^{x}, \mathbf{c}^{x}, \mathbf{s}_{\phi}^{x}, \boldsymbol{\omega}^{x}\},
  \qquad
  \mathcal{D}(\bP)
  =
  \{\mathbf{a}^{p}, \mathbf{c}^{p}, \mathbf{s}_{\phi}^{p}, \boldsymbol{\omega}^{p}\}.
\end{equation}
These descriptors instantiate the local cycle state used by \Method: the log-amplitude descriptor captures cycle-magnitude variation, the phase sine/cosine descriptors capture cycle-alignment variation, and the phase-increment descriptor provides cues about local cycle-duration variation. They therefore connect the NRP motivation to the gate input without requiring a predefined period or a hard template index.

\subsubsection{Descriptor-Conditioned Gate}
For each channel, we concatenate the descriptors of the observation and the prior along the feature dimension:
\begin{equation}
  \mathbf{u}_{c}
  =
  \operatorname{Concat}\!\left[
    \mathbf{a}^{x}_{c},
    \mathbf{a}^{p}_{c},
    \mathbf{c}^{x}_{c},
    \mathbf{s}_{\phi,c}^{x},
    \mathbf{c}^{p}_{c},
    \mathbf{s}_{\phi,c}^{p},
    \boldsymbol{\omega}^{x}_{c},
    \boldsymbol{\omega}^{p}_{c}
    \right]
  \in \R^{8 \times L}.
\end{equation}
A lightweight one-dimensional convolutional gate then produces a local alignment coefficient:
\begin{equation}
  \mathbf{g}_{c}
  =
  \sigma\!\left(\operatorname{Conv}_{2}\!\left(
    \phi\!\left(\operatorname{Conv}_{1}(\mathbf{u}_{c})\right)
    \right)\right),
  \qquad
  \mathbf{g}_{c} \in [0,1]^L,
\end{equation}
where $\phi(\cdot)$ is the GELU (Gaussian Error Linear Unit) activation~\cite{Hendrycks-2016-GELU} and $\sigma(\cdot)$ is the sigmoid function. Because the gate is conditioned on amplitude, phase, and local frequency cues from both $\bX$ and $\bP$, it can learn where the observation should be trusted and where the oscillatory reference should contribute more strongly.

\subsubsection{State Alignment}
The gate performs local oscillatory-state alignment by adaptively interpolating between the observation and the global oscillatory prior:
\begin{equation}
  \widetilde{\bX}
  =
  (1-\mathbf{G}) \odot \bX + \mathbf{G} \odot \bP,
  \label{eq:hilbert_correction}
\end{equation}
where $\mathbf{G}\in[0,1]^{C \times L}$ stacks all channel gates. The gate can preserve the raw sequence where the observation is reliable and increase the contribution of $\bP$ where the descriptor comparison indicates a local mismatch. This mechanism provides an explicit temporal inductive bias through gated residual alignment rather than fixed period indexing or hard template matching.

\subsection{Dual-Path Forecasting and Fusion}

After oscillatory-state alignment, \Method feeds the aligned sequence $\widetilde{\bX}\in\R^{C\times L}$ into two complementary forecasting paths (see Fig.~\ref{fig:framework}). The base path provides a stable per-channel extrapolation anchor, while the attention path models cross-variate nonlinear corrections after alignment. This separation keeps the main temporal inductive bias in the AOS module rather than absorbing it into a more complex backbone.

\subsubsection{Base Path}
This path provides the stable per-channel extrapolation anchor. For each channel, a shared weight matrix directly maps the aligned look-back window to the prediction horizon:
\begin{equation}
  \widehat{\by}_{\mathrm{base},c}
  =
  \mathbf{W}_{l}\widetilde{\bx}_{c}+\mathbf{b}_{l},
  \qquad
  \mathbf{W}_{l}\in\R^{H\times L}.
\end{equation}
Stacking all channels gives $\widehat{\bY}_{\mathrm{base}}\in\R^{C\times H}$.

\subsubsection{Attention Path}
This path refines cross-variate dependencies after oscillatory-state alignment. A shared temporal embedding first projects each channel from the look-back length $L$ into a hidden dimension $d$, with dropout regularization~\cite{Srivastava-2014-Dropout}:
\begin{equation}
  \mathbf{h}_{c}
  =
  \operatorname{Dropout}\!\left(
  \phi(\mathbf{W}_{e}\widetilde{\bx}_{c}+\mathbf{b}_{e})
  \right),
  \qquad
  \mathbf{h}_{c} \in \R^{d}.
\end{equation}
Stacking all channels gives $\mathbf{H}\in\R^{C \times d}$. A residual multi-head self-attention (MHA) layer~\cite{Vaswani-2017-transformer} then refines cross-variate interactions over these channel tokens:
\begin{equation}
  \mathbf{R}
  =
  \operatorname{MHA}(\mathbf{H},\mathbf{H},\mathbf{H}) + \mathbf{H}.
\end{equation}
The refined features are passed through a shallow nonlinear projection and output head:
\begin{equation}
  \widehat{\bY}_{\mathrm{attn}}
  =
  \mathbf{W}_{o}\,\phi(\mathbf{W}_{r}\mathbf{R}+\mathbf{b}_{r})+\mathbf{b}_{o}.
\end{equation}

\subsubsection{Prediction Fusion}
The final prediction is a learnable combination of the two paths:
\begin{equation}
  \widehat{\bY}
  =
  \lambda\,\widehat{\bY}_{\mathrm{attn}}
  +
  (1-\lambda)\,\widehat{\bY}_{\mathrm{base}},
  \qquad
  \lambda=\sigma(\eta),
  \label{eq:output_fusion}
\end{equation}
where $\eta$ is a scalar learnable parameter. This dual-path design lets the model adaptively balance stable channel-wise extrapolation from the base path and nonlinear cross-variate correction from the attention path.

\section{Experiments}
\label{sec:experiments}

In this section, we evaluate \Method by answering the following research questions:
\begin{itemize}
  \item \textbf{RQ1}: Does \Method achieve competitive forecasting accuracy on standard benchmarks? (Sec.~\ref{sec:main_results})
  \item \textbf{RQ2}: Does \Method provide practical gains for cloud workload forecasting with efficient inference? (Sec.~\ref{sec:workload_efficiency})
  \item \textbf{RQ3}: Does adaptive oscillatory-state alignment overcome fixed-template failure under controlled NRP? (Sec.~\ref{sec:failure_modes})
  \item \textbf{RQ4}: Which components and settings drive the performance? (Sec.~\ref{sec:ablation})
  \item \textbf{RQ5}: Do the learned prior and gate implement oscillatory-state alignment? (Sec.~\ref{sec:alignment_analysis})
\end{itemize}

\subsection{Setup}
\label{sec:experimental_setup}

All experiments are implemented in PyTorch~\cite{Paszke-2019-Pytorch}. We evaluate forecasting accuracy with mean squared error (MSE) and mean absolute error (MAE), where lower values indicate better performance. Following the standard long-term forecasting protocol, the look-back length is fixed to $L=96$ for all benchmark comparisons.

\subsubsection{Datasets}
We evaluate on eight widely used multivariate forecasting benchmarks: ETTh1, ETTh2, ETTm1, ETTm2 from the ETT series~\cite{Zhou-2021-Informerefficienttransformer}; Electricity, Solar-Energy, Traffic, and Weather~\cite{Wu-2021-Autoformer}. We further evaluate on two cloud workload traces (IaaS and PaaS) sampled at 10-minute intervals. For all datasets, the prediction horizons are $H\in\{96,192,336,720\}$ with a fixed look-back length $L=96$. Detailed dataset statistics are summarized in Table~\ref{tab:dataset}.

\begin{table}[t]
  \centering
  \caption{Detailed information about the datasets used in this study.}
  \label{tab:dataset}
  \begin{adjustbox}{max width=\linewidth}
    \begin{tabular}{@{}lcccc@{}}
      \toprule
      Dataset     & Channels & Timesteps & Interval & Domain         \\ \midrule
      \multicolumn{5}{l}{\textit{Common Benchmarks}}                 \\
      ETTh1       & 7        & 14,400    & 1 hour   & Electricity    \\
      ETTh2       & 7        & 14,400    & 1 hour   & Electricity    \\
      ETTm1       & 7        & 57,600    & 15 mins  & Electricity    \\
      ETTm2       & 7        & 57,600    & 15 mins  & Electricity    \\
      Electricity & 321      & 26,304    & 1 hour   & Electricity    \\
      Solar       & 137      & 52,560    & 10 mins  & Energy         \\
      Traffic     & 862      & 17,544    & 1 hour   & Transportation \\
      Weather     & 21       & 52,696    & 10 mins  & Weather        \\
      \midrule
      \multicolumn{5}{l}{\textit{Workload Traces}}                   \\
      IaaS        & 93       & 3,456     & 10 mins  & Cloud workload \\
      PaaS        & 426      & 7,776     & 10 mins  & Cloud workload \\ \bottomrule
    \end{tabular}
  \end{adjustbox}
\end{table}

\subsubsection{Baselines}
We compare \Method with representative recent forecasting models, including TQNet~\cite{Lin-2025-TQNet}, SRSNet~\cite{Wu-2025-SRSNet}, SSformer~\cite{SSformer-2026-AAAI}, CycleNet~\cite{Lin-2024-CycleNet}, Amplifier~\cite{Fei-2025-Amplifier}, iTransformer~\cite{Liu-2024-iTransformer}, PatchTST~\cite{Nie-2023-PacthTST}, and DLinear~\cite{Zeng-2023-DLinear}. We use TQNet's reported results when available, take SSformer results from the original paper, and keep the same benchmark protocol for all compared methods.

\subsection{RQ1: Forecasting Accuracy on Public Benchmarks}
\label{sec:main_results}

\begin{table*}[t]
  \centering
  \scriptsize
  \caption{Full multivariate forecasting results for all prediction horizons. The look-back length is fixed to $L=96$. 
  \Method values are mean $\pm$ std over 5 runs. }
  \label{tab:main_results}
  \begin{adjustbox}{max width=\textwidth}
    \setlength{\tabcolsep}{5pt}
    \begin{tabular}{@{}cc|cc|cc|cc|cc|cc|cc|cc|cc|cc@{}}
      \toprule
      \multicolumn{2}{c|}{Model}                    & \multicolumn{2}{c|}{\Method} & \multicolumn{2}{c|}{TQNet}  & \multicolumn{2}{c|}{SRSNet} & \multicolumn{2}{c|}{SSformer} & \multicolumn{2}{c|}{CycleNet} & \multicolumn{2}{c|}{Amplifier} & \multicolumn{2}{c|}{iTransformer} & \multicolumn{2}{c|}{PatchTST} & \multicolumn{2}{c}{DLinear}                                                                                                                                   \\
      \multicolumn{2}{c|}{}                         & \multicolumn{2}{c|}{(Ours)}  & \multicolumn{2}{c|}{(2025)} & \multicolumn{2}{c|}{(2025)} & \multicolumn{2}{c|}{(2026)}   & \multicolumn{2}{c|}{(2024)}   & \multicolumn{2}{c|}{(2025)}    & \multicolumn{2}{c|}{(2024)}       & \multicolumn{2}{c|}{(2023)}   & \multicolumn{2}{c}{(2023)}                                                                                                                                    \\
      \multicolumn{2}{c|}{Metric}                   & MSE                          & MAE                         & MSE                         & MAE                           & MSE                           & MAE                            & MSE                               & MAE                           & MSE                         & MAE             & MSE             & MAE   & MSE             & MAE             & MSE             & MAE   & MSE   & MAE           \\
      \midrule
      \multirow{5}{*}{\rotatebox{90}{ETTh1}}        & 96                           & \rank{1}{0.368\std{0.002}}  & \rank{1}{0.392\std{0.001}}  & \rank{2}{0.371}               & \rank{2}{0.393}               & 0.383                          & 0.395                             & 0.373                         & 0.393                       & 0.375           & 0.395           & 0.376 & 0.393           & 0.386           & 0.405           & 0.414 & 0.419 & 0.386 & 0.400 \\
                                                    & 192                          & \rank{1}{0.421\std{0.003}}  & \rank{1}{0.421\std{0.002}}  & \rank{2}{0.428}               & 0.426                         & 0.433                          & \rank{2}{0.422}                   & 0.432                         & 0.431                       & 0.436           & 0.428           & 0.442 & 0.430           & 0.441           & 0.436           & 0.460 & 0.445 & 0.437 & 0.432 \\
                                                    & 336                          & \rank{1}{0.459\std{0.002}}  & \rank{1}{0.440\std{0.003}}  & 0.476                         & \rank{2}{0.446}               & 0.476                          & 0.446                             & 0.483                         & 0.454                       & 0.496           & 0.455           & 0.478 & 0.446           & 0.487           & 0.458           & 0.501 & 0.466 & 0.481 & 0.459 \\
                                                    & 720                          & \rank{1}{0.462\std{0.003}}  & \rank{1}{0.461\std{0.002}}  & 0.487                         & \rank{2}{0.470}               & \rank{2}{0.474}                & 0.471                             & 0.519                         & 0.495                       & 0.520           & 0.484           & 0.501 & 0.479           & 0.503           & 0.491           & 0.500 & 0.488 & 0.519 & 0.516 \\
      \cmidrule(lr){3-20}
                                                    & Avg                          & \rank{1}{0.427\std{0.003}}  & \rank{1}{0.429\std{0.002}}  & \rank{2}{0.441}               & 0.434                         & 0.442                          & \rank{2}{0.433}                   & 0.452                         & 0.443                       & 0.457           & 0.441           & 0.449 & 0.437           & 0.454           & 0.448           & 0.469 & 0.455 & 0.456 & 0.452 \\
      \midrule
      \multirow{5}{*}{\rotatebox{90}{ETTh2}}        & 96                           & \rank{1}{0.280\std{0.002}}  & \rank{1}{0.329\std{0.002}}  & 0.295                         & 0.343                         & 0.296                          & 0.345                             & \rank{2}{0.281}               & \rank{2}{0.332}             & 0.298           & 0.344           & 0.298 & 0.347           & 0.297           & 0.349           & 0.302 & 0.348 & 0.333 & 0.387 \\
                                                    & 192                          & \rank{1}{0.354\std{0.003}}  & \rank{1}{0.378\std{0.001}}  & \rank{2}{0.367}               & 0.393                         & 0.369                          & 0.392                             & 0.367                         & \rank{2}{0.385}             & 0.372           & 0.396           & 0.378 & 0.401           & 0.380           & 0.400           & 0.388 & 0.400 & 0.477 & 0.476 \\
                                                    & 336                          & \rank{2}{0.396\std{0.001}}  & \rank{1}{0.414\std{0.002}}  & 0.417                         & 0.427                         & 0.413                          & 0.425                             & \rank{1}{0.394}               & \rank{2}{0.418}             & 0.431           & 0.439           & 0.428 & 0.437           & 0.428           & 0.432           & 0.426 & 0.433 & 0.594 & 0.541 \\
                                                    & 720                          & \rank{2}{0.410\std{0.002}}  & \rank{2}{0.428\std{0.003}}  & 0.433                         & 0.446                         & 0.425                          & 0.444                             & \rank{1}{0.395}               & \rank{1}{0.420}             & 0.450           & 0.458           & 0.452 & 0.460           & 0.427           & 0.445           & 0.431 & 0.446 & 0.831 & 0.657 \\
      \cmidrule(lr){3-20}
                                                    & Avg                          & \rank{2}{0.360\std{0.002}}  & \rank{1}{0.387\std{0.002}}  & 0.378                         & 0.402                         & 0.376                          & 0.402                             & \rank{1}{0.359}               & \rank{2}{0.389}             & 0.388           & 0.409           & 0.389 & 0.411           & 0.383           & 0.407           & 0.387 & 0.407 & 0.559 & 0.515 \\
      \midrule
      \multirow{5}{*}{\rotatebox{90}{ETTm1}}        & 96                           & \rank{1}{0.307\std{0.001}}  & \rank{1}{0.345\std{0.002}}  & \rank{2}{0.311}               & \rank{2}{0.353}               & 0.319                          & 0.358                             & 0.318                         & 0.355                       & 0.319           & 0.360           & 0.318 & 0.356           & 0.334           & 0.368           & 0.329 & 0.367 & 0.345 & 0.372 \\
                                                    & 192                          & \rank{1}{0.356\std{0.002}}  & \rank{1}{0.375\std{0.001}}  & \rank{2}{0.356}               & \rank{2}{0.378}               & 0.359                          & 0.381                             & 0.360                         & 0.379                       & 0.360           & 0.381           & 0.362 & 0.381           & 0.377           & 0.391           & 0.367 & 0.385 & 0.380 & 0.389 \\
                                                    & 336                          & 0.391\std{0.003}            & \rank{1}{0.399\std{0.002}}  & \rank{2}{0.390}               & \rank{2}{0.401}               & 0.391                          & 0.404                             & 0.393                         & 0.402                       & \rank{1}{0.389} & 0.403           & 0.393 & 0.404           & 0.426           & 0.420           & 0.399 & 0.410 & 0.413 & 0.413 \\
                                                    & 720                          & 0.452\std{0.002}            & \rank{2}{0.437\std{0.003}}  & 0.452                         & 0.440                         & 0.470                          & \rank{1}{0.436}                   & 0.449                         & 0.440                       & \rank{1}{0.447} & 0.441           & 0.460 & 0.442           & 0.491           & 0.459           & 0.454 & 0.439 & 0.474 & 0.453 \\
      \cmidrule(lr){3-20}
                                                    & Avg                          & \rank{1}{0.377\std{0.002}}  & \rank{1}{0.389\std{0.002}}  & \rank{1}{0.377}               & \rank{2}{0.393}               & 0.385                          & 0.395                             & 0.380                         & 0.394                       & 0.379           & 0.396           & 0.383 & 0.396           & 0.407           & 0.410           & 0.387 & 0.400 & 0.403 & 0.407 \\
      \midrule
      \multirow{5}{*}{\rotatebox{90}{ETTm2}}        & 96                           & \rank{2}{0.168\std{0.001}}  & \rank{2}{0.247\std{0.002}}  & 0.173                         & 0.256                         & 0.181                          & 0.267                             & 0.171                         & 0.254                       & \rank{1}{0.163} & \rank{1}{0.246} & 0.178 & 0.261           & 0.180           & 0.264           & 0.175 & 0.259 & 0.193 & 0.292 \\
                                                    & 192                          & \rank{2}{0.233\std{0.002}}  & \rank{2}{0.291\std{0.001}}  & 0.238                         & 0.298                         & 0.243                          & 0.306                             & 0.235                         & 0.302                       & \rank{1}{0.229} & \rank{1}{0.290} & 0.243 & 0.303           & 0.250           & 0.309           & 0.241 & 0.302 & 0.284 & 0.362 \\
                                                    & 336                          & \rank{2}{0.289\std{0.003}}  & \rank{1}{0.326\std{0.002}}  & 0.301                         & 0.340                         & 0.306                          & 0.346                             & 0.307                         & 0.338                       & \rank{1}{0.284} & \rank{2}{0.327} & 0.305 & 0.344           & 0.311           & 0.348           & 0.305 & 0.343 & 0.369 & 0.427 \\
                                                    & 720                          & \rank{1}{0.384\std{0.002}}  & \rank{1}{0.385\std{0.003}}  & 0.397                         & 0.396                         & 0.407                          & 0.399                             & 0.403                         & 0.395                       & \rank{2}{0.389} & \rank{2}{0.391} & 0.393 & 0.397           & 0.412           & 0.407           & 0.402 & 0.400 & 0.554 & 0.522 \\
      \cmidrule(lr){3-20}
                                                    & Avg                          & \rank{2}{0.269\std{0.002}}  & \rank{1}{0.312\std{0.002}}  & 0.277                         & 0.323                         & 0.284                          & 0.329                             & 0.279                         & 0.322                       & \rank{1}{0.266} & \rank{2}{0.314} & 0.280 & 0.326           & 0.288           & 0.332           & 0.281 & 0.326 & 0.350 & 0.401 \\
      \midrule
      \multirow{5}{*}{\rotatebox{90}{Electricity}}  & 96                           & \rank{1}{0.132\std{0.001}}  & \rank{1}{0.225\std{0.001}}  & \rank{2}{0.134}               & \rank{2}{0.229}               & 0.161                          & 0.252                             & 0.141                         & 0.238                       & 0.136           & 0.229           & 0.149 & 0.245           & 0.148           & 0.240           & 0.181 & 0.270 & 0.197 & 0.282 \\
                                                    & 192                          & \rank{1}{0.150\std{0.002}}  & \rank{1}{0.242\std{0.001}}  & 0.154                         & 0.247                         & 0.172                          & 0.261                             & 0.157                         & 0.251                       & \rank{2}{0.152} & \rank{2}{0.244} & 0.165 & 0.260           & 0.162           & 0.253           & 0.188 & 0.274 & 0.196 & 0.285 \\
                                                    & 336                          & \rank{2}{0.170\std{0.001}}  & \rank{1}{0.264\std{0.002}}  & \rank{1}{0.169}               & \rank{1}{0.264}               & 0.190                          & 0.279                             & 0.174                         & 0.268                       & \rank{2}{0.170} & \rank{1}{0.264} & 0.176 & 0.271           & 0.178           & 0.269           & 0.204 & 0.293 & 0.209 & 0.301 \\
                                                    & 720                          & \rank{1}{0.184\std{0.002}}  & \rank{1}{0.278\std{0.001}}  & \rank{2}{0.201}               & \rank{2}{0.294}               & 0.231                          & 0.313                             & 0.209                         & 0.303                       & 0.212           & 0.299           & 0.204 & 0.296           & 0.225           & 0.317           & 0.246 & 0.324 & 0.245 & 0.333 \\
      \cmidrule(lr){3-20}
                                                    & Avg                          & \rank{1}{0.159\std{0.002}}  & \rank{1}{0.252\std{0.001}}  & \rank{2}{0.164}               & \rank{2}{0.259}               & 0.189                          & 0.276                             & 0.170                         & 0.265                       & 0.168           & \rank{2}{0.259} & 0.174 & 0.268           & 0.178           & 0.270           & 0.205 & 0.290 & 0.212 & 0.300 \\
      \midrule
      \multirow{5}{*}{\rotatebox{90}{Solar-Energy}} & 96                           & \rank{2}{0.187\std{0.002}}  & \rank{1}{0.215\std{0.001}}  & \rank{1}{0.173}               & 0.233                         & 0.216                          & 0.258                             & 0.197                         & 0.236                       & 0.190           & 0.247           & 0.186 & \rank{2}{0.232} & 0.203           & 0.237           & 0.234 & 0.286 & 0.290 & 0.378 \\
                                                    & 192                          & 0.221\std{0.001}            & \rank{1}{0.249\std{0.002}}  & \rank{1}{0.199}               & 0.257                         & 0.247                          & 0.280                             & 0.222                         & 0.268                       & \rank{2}{0.210} & \rank{2}{0.266} & 0.231 & 0.264           & 0.233           & 0.261           & 0.267 & 0.310 & 0.320 & 0.398 \\
                                                    & 336                          & 0.234\std{0.002}            & \rank{2}{0.262\std{0.001}}  & \rank{1}{0.211}               & \rank{2}{0.263}               & 0.268                          & 0.294                             & 0.245                         & 0.271                       & \rank{2}{0.217} & 0.266           & 0.234 & 0.263           & 0.248           & 0.273           & 0.290 & 0.315 & 0.353 & 0.415 \\
                                                    & 720                          & 0.244\std{0.001}            & 0.271\std{0.002}            & \rank{1}{0.209}               & 0.270                         & 0.268                          & 0.290                             & 0.247                         & 0.287                       & \rank{2}{0.223} & \rank{2}{0.266} & 0.238 & \rank{1}{0.265} & 0.249           & 0.275           & 0.289 & 0.317 & 0.356 & 0.413 \\
      \cmidrule(lr){3-20}
                                                    & Avg                          & 0.222\std{0.002}            & \rank{1}{0.249\std{0.002}}  & \rank{1}{0.198}               & \rank{2}{0.256}               & 0.250                          & 0.281                             & 0.228                         & 0.266                       & \rank{2}{0.210} & 0.261           & 0.222 & 0.256           & 0.233           & 0.262           & 0.270 & 0.307 & 0.330 & 0.401 \\
      \midrule
      \multirow{5}{*}{\rotatebox{90}{Traffic}}      & 96                           & \rank{2}{0.411\std{0.002}}  & \rank{2}{0.267\std{0.001}}  & 0.413                         & \rank{1}{0.261}               & 0.471                          & 0.295                             & 0.420                         & 0.288                       & 0.458           & 0.296           & 0.450 & 0.295           & \rank{1}{0.395} & 0.268           & 0.462 & 0.290 & 0.650 & 0.396 \\
                                                    & 192                          & \rank{2}{0.429\std{0.001}}  & \rank{2}{0.276\std{0.002}}  & 0.432                         & \rank{1}{0.271}               & 0.480                          & 0.300                             & 0.434                         & 0.298                       & 0.457           & 0.294           & 0.489 & 0.311           & \rank{1}{0.417} & \rank{2}{0.276} & 0.466 & 0.290 & 0.598 & 0.370 \\
                                                    & 336                          & 0.459\std{0.003}            & 0.286\std{0.001}            & \rank{2}{0.450}               & \rank{1}{0.277}               & 0.496                          & 0.306                             & 0.468                         & 0.301                       & 0.470           & 0.299           & 0.484 & 0.321           & \rank{1}{0.433} & \rank{2}{0.283} & 0.482 & 0.300 & 0.605 & 0.373 \\
                                                    & 720                          & \rank{2}{0.480\std{0.002}}  & 0.304\std{0.002}            & \rank{2}{0.486}               & \rank{1}{0.295}               & 0.531                          & 0.328                             & 0.487                         & 0.309                       & 0.502           & 0.314           & 0.517 & 0.333           & \rank{1}{0.467} & \rank{2}{0.302} & 0.514 & 0.320 & 0.645 & 0.394 \\
      \cmidrule(lr){3-20}
                                                    & Avg                          & \rank{2}{0.445\std{0.002}}  & 0.283\std{0.002}            & \rank{2}{0.445}               & \rank{1}{0.276}               & 0.494                          & 0.307                             & 0.452                         & 0.299                       & 0.472           & 0.301           & 0.485 & 0.315           & \rank{1}{0.428} & \rank{2}{0.282} & 0.481 & 0.300 & 0.625 & 0.383 \\
      \midrule
      \multirow{5}{*}{\rotatebox{90}{Weather}}      & 96                           & \rank{1}{0.148\std{0.001}}  & \rank{1}{0.189\std{0.001}}  & 0.157                         & \rank{2}{0.200}               & 0.167                          & 0.214                             & \rank{2}{0.156}               & \rank{2}{0.204}             & 0.158           & 0.203           & 0.165 & 0.210           & 0.174           & 0.214           & 0.177 & 0.210 & 0.196 & 0.255 \\
                                                    & 192                          & \rank{1}{0.200\std{0.002}}  & \rank{1}{0.241\std{0.001}}  & 0.206                         & \rank{2}{0.245}               & 0.215                          & 0.255                             & \rank{2}{0.203}               & \rank{2}{0.245}             & 0.207           & 0.247           & 0.212 & 0.253           & 0.221           & 0.254           & 0.225 & 0.250 & 0.237 & 0.296 \\
                                                    & 336                          & \rank{2}{0.256\std{0.001}}  & \rank{1}{0.284\std{0.002}}  & 0.262                         & 0.287                         & 0.270                          & 0.294                             & \rank{1}{0.250}               & \rank{1}{0.284}             & 0.262           & 0.289           & 0.267 & 0.293           & 0.278           & 0.296           & 0.278 & 0.290 & 0.283 & 0.335 \\
                                                    & 720                          & \rank{1}{0.336\std{0.002}}  & \rank{2}{0.339\std{0.001}}  & 0.344                         & 0.342                         & 0.346                          & 0.344                             & \rank{2}{0.338}               & \rank{1}{0.337}             & 0.344           & 0.344           & 0.344 & 0.342           & 0.358           & 0.349           & 0.354 & 0.340 & 0.345 & 0.381 \\
      \cmidrule(lr){3-20}
                                                    & Avg                          & \rank{1}{0.235\std{0.002}}  & \rank{1}{0.263\std{0.001}}  & 0.242                         & 0.269                         & 0.250                          & 0.277                             & \rank{2}{0.237}               & \rank{2}{0.268}             & 0.243           & 0.271           & 0.247 & 0.275           & 0.258           & 0.278           & 0.259 & 0.273 & 0.265 & 0.317 \\
      \bottomrule
    \end{tabular}
  \end{adjustbox}
\end{table*}

\subsubsection{Common Benchmarks}

Table~\ref{tab:main_results} reports the full results across all horizons. \Method achieves the best or second-best MSE on $7$ out of $8$ datasets and the best MAE on $7$ out of $8$ datasets; the only exceptions are Solar-Energy (where the best MSE belongs to TQNet) and Traffic (where the best MAE belongs to TQNet). Compared with the most relevant fixed-template periodic-prior baselines (TQNet, CycleNet), \Method consistently improves accuracy, particularly on datasets with complex oscillatory dynamics such as Electricity ($-3.0\%$ MSE vs.\ TQNet) and ETTh2 ($-4.8\%$ MSE vs.\ TQNet). These gains are consistent with our motivation: adaptive oscillatory-state alignment is most beneficial when recurrence persists but local cycle states---magnitude, alignment, or duration---evolve over time.

\subsubsection{Stability}
To assess statistical significance, we report all \Method results as the mean and standard deviation over five independent runs with different random seeds. As shown in Table~\ref{tab:main_results}, the standard deviations remain small across all datasets and horizons---MSE std is at most $0.003$ and MAE std at most $0.003$---indicating that \Method has low variance across random initializations.

\subsection{RQ2: Cloud Workload Forecasting and Efficiency}
\label{sec:workload_efficiency}

\subsubsection{Workload Forecasting}
\label{sec:workload_results}
Accurate workload forecasting is central to capacity planning, autoscaling, and cloud resource management. We therefore evaluate on two cloud workload traces (IaaS and PaaS) sampled every 10 minutes. In addition to MSE and MAE, we report normalized MAE (NMAE). Table~\ref{tab:workload_results} presents the full per-horizon results.
\Method ranks first on all metrics for IaaS and on average for PaaS, with the advantage over TQNet growing at longer horizons (MSE gap from $0.000$ at $H{=}96$ to $0.043$ at $H{=}720$ on IaaS). These results indicate that adaptive oscillatory-state alignment transfers effectively to operational cloud workload forecasting.

\begin{figure}[t]
  \centering
  \includegraphics[width=\linewidth]{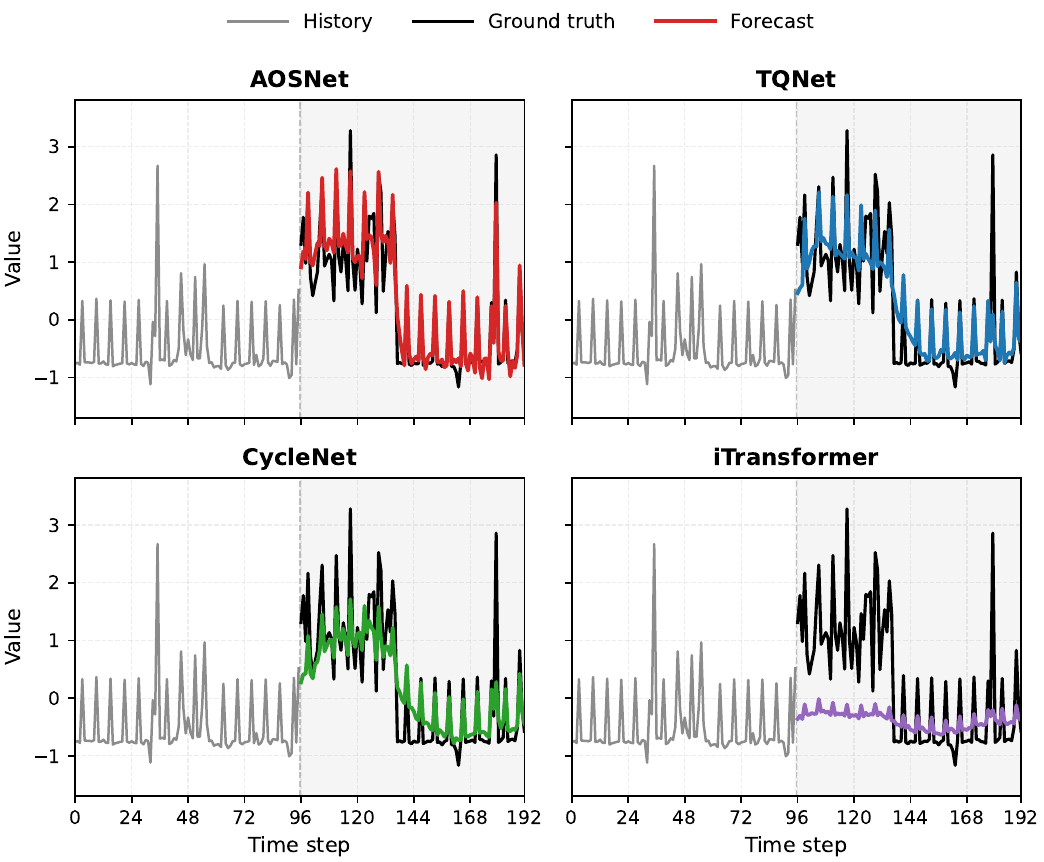}
  \caption{Representative IaaS 96$\rightarrow$96 forecasting visualization on a single channel. Each panel compares the same history window (gray), ground truth (black), and model forecast (colored) for \Method, TQNet, CycleNet, and iTransformer.}
  \label{fig:iaas_case_study}
\end{figure}

Fig.~\ref{fig:iaas_case_study} shows a representative IaaS 96$\rightarrow$96 case. On this sample, \Method follows the post-history regime transition more faithfully, while TQNet and CycleNet remain closer to rigid template-like dynamics and iTransformer underfits the oscillatory pattern.

\begin{table}[t]
  \centering
  \scriptsize
  \caption{Workload forecasting results for all prediction horizons with fixed look-back $L=96$. Lower is better. Best results are in \textbf{bold}, second best are \underline{underlined}.}
  \label{tab:workload_results}
  \begin{adjustbox}{max width=\linewidth}
    \setlength{\tabcolsep}{3pt}
    \begin{tabular}{@{}cc|ccc|ccc|ccc|ccc@{}}
      \toprule
      \multicolumn{2}{c|}{Dataset / $H$} & \multicolumn{3}{c|}{\Method (Ours)} & \multicolumn{3}{c|}{TQNet} & \multicolumn{3}{c|}{CycleNet} & \multicolumn{3}{c}{iTransformer}                                                                                                                                     \\
                                         &                                     & MSE                        & MAE                           & NMAE                             & MSE             & MAE             & NMAE            & MSE             & MAE             & NMAE            & MSE   & MAE   & NMAE  \\
      \midrule
      \multirow{5}{*}{\rotatebox{90}{IaaS}}
                                         & 96                                  & \rank{1}{0.754}            & \rank{1}{0.596}               & \rank{1}{0.748}                  & \rank{2}{0.754} & \rank{2}{0.602} & \rank{2}{0.756} & 0.756           & 0.604           & 0.759           & 0.905 & 0.684 & 0.859 \\
                                         & 192                                 & \rank{1}{0.723}            & \rank{1}{0.583}               & \rank{1}{0.716}                  & \rank{2}{0.740} & \rank{2}{0.588} & \rank{2}{0.722} & 0.754           & 0.591           & 0.725           & 0.799 & 0.621 & 0.763 \\
                                         & 336                                 & \rank{1}{0.726}            & \rank{1}{0.577}               & \rank{1}{0.701}                  & 0.746           & 0.592           & 0.719           & \rank{2}{0.743} & \rank{2}{0.587} & \rank{2}{0.714} & 0.810 & 0.627 & 0.762 \\
                                         & 720                                 & \rank{1}{0.805}            & \rank{1}{0.608}               & \rank{1}{0.714}                  & \rank{2}{0.848} & \rank{2}{0.630} & \rank{2}{0.741} & 0.906           & 0.660           & 0.775           & 1.025 & 0.718 & 0.843 \\
                                         & Avg                                 & \rank{1}{0.752}            & \rank{1}{0.591}               & \rank{1}{0.720}                  & \rank{2}{0.772} & \rank{2}{0.603} & \rank{2}{0.735} & 0.790           & 0.610           & 0.743           & 0.885 & 0.662 & 0.807 \\
      \midrule
      \multirow{5}{*}{\rotatebox{90}{PaaS}}
                                         & 96                                  & 0.076                      & \rank{2}{0.144}               & \rank{2}{0.186}                  & \rank{1}{0.062} & \rank{1}{0.138} & \rank{1}{0.180} & \rank{2}{0.074} & 0.142           & 0.185           & 0.196 & 0.272 & 0.354 \\
                                         & 192                                 & \rank{1}{0.139}            & \rank{1}{0.170}               & \rank{1}{0.181}                  & \rank{2}{0.149} & \rank{2}{0.179} & \rank{2}{0.192} & 0.145           & 0.178           & 0.191           & 0.346 & 0.341 & 0.365 \\
                                         & 336                                 & \rank{1}{0.136}            & 0.180                         & \rank{1}{0.195}                  & 0.148           & 0.194           & 0.209           & \rank{2}{0.136} & \rank{1}{0.173} & \rank{2}{0.187} & 0.351 & 0.340 & 0.368 \\
                                         & 720                                 & \rank{2}{0.141}            & \rank{1}{0.221}               & \rank{1}{0.240}                  & \rank{1}{0.136} & \rank{2}{0.223} & \rank{2}{0.242} & 0.166           & 0.255           & 0.278           & 0.455 & 0.420 & 0.457 \\
                                         & Avg                                 & \rank{1}{0.123}            & \rank{1}{0.178}               & \rank{1}{0.201}                  & \rank{2}{0.124} & \rank{2}{0.183} & \rank{2}{0.206} & 0.130           & 0.187           & 0.210           & 0.337 & 0.343 & 0.386 \\
      \bottomrule
    \end{tabular}
  \end{adjustbox}
\end{table}

\begin{figure}[t]
  \centering
  \includegraphics[width=\linewidth]{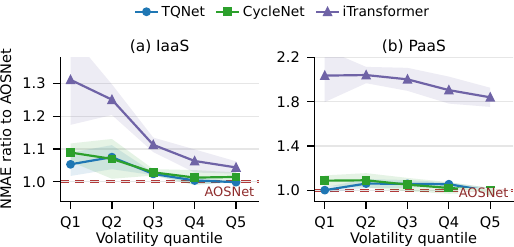}
  \caption{Volatility-stratified workload forecasting errors. Test windows are divided into five quantile bins by target-window volatility within each dataset and horizon. Curves report the mean NMAE ratio to \Method over horizons $H\in\{96,192,336,720\}$; shaded bands indicate the standard error across horizons. The dashed line denotes \Method.}
  \label{fig:workload_stratified}
\end{figure}

To examine whether the workload gains are concentrated in a narrow low-volatility subset, we further stratify the test windows by target-window volatility. For each dataset and horizon, we split windows into five quantile bins according to the mean absolute first difference normalized by the target-window standard deviation, and then aggregate the NMAE ratio to \Method across all four horizons. Fig.~\ref{fig:workload_stratified} shows that fixed-template periodic baselines are usually above the \Method reference line across the volatility spectrum, while iTransformer incurs consistently larger errors. This suggests that the workload benefit is not driven by a small low-volatility subset; at the same time, the highest-volatility bins show that local volatility alone does not fully explain all remaining errors, motivating the complementary horizon-wise results in Table~\ref{tab:workload_results}.

\subsubsection{Efficiency}
\label{sec:efficiency}

\begin{figure*}[t]
  \centering
  \begin{minipage}{0.45\linewidth}
    \centering
    \includegraphics[width=\linewidth]{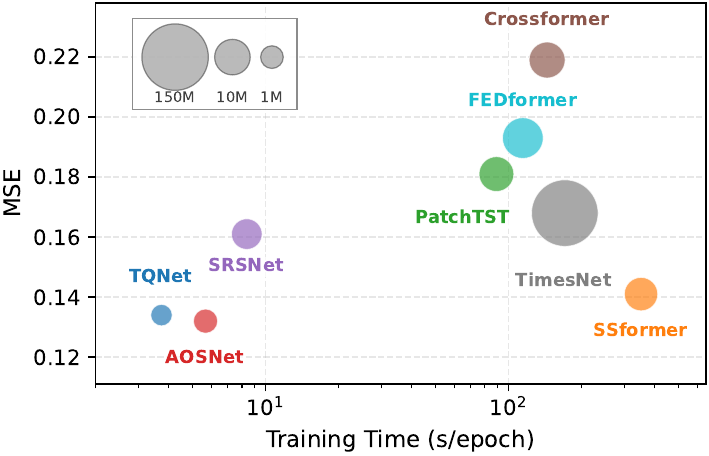}
  \end{minipage}
  \hfill
  \begin{minipage}{0.45\linewidth}
    \centering
    \includegraphics[width=\linewidth]{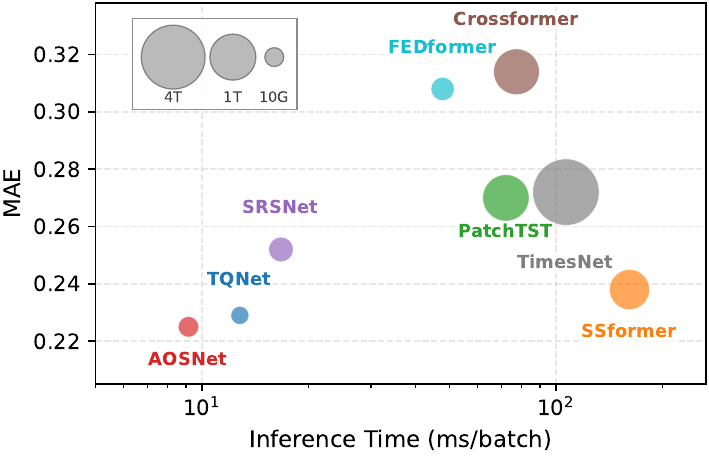}
  \end{minipage}
  \caption{Efficiency comparison on Electricity (look-back $96$, horizon $96$, batch size $32$). Left: MSE vs.\ training time per epoch, with bubble size encoding parameters. Right: MAE vs.\ inference time per batch, with bubble size encoding FLOPs. Parameter bubbles are reported in millions, while FLOP bubbles use the units shown in the inset.}
  \label{fig:efficiency}
\end{figure*}

Fig.~\ref{fig:efficiency} shows that \Method achieves the best accuracy--efficiency trade-off among the evaluated models. On Electricity 96$\rightarrow$96, it obtains the lowest MSE/MAE with the fastest inference (9.15\,ms/batch)---$1.4{\times}$ faster than TQNet, $7.9{\times}$ faster than PatchTST, $17.6{\times}$ faster than SSformer, $8.4{\times}$ faster than Crossformer, $5.2{\times}$ faster than FEDformer, and $11.7{\times}$ faster than TimesNet. \Method also remains compact with 1.45M parameters, compared with 10.60M for Crossformer, 18.44M for FEDformer, and 150.30M for TimesNet. This efficiency is important for cloud workload forecasting, where prediction models may be invoked repeatedly in capacity planning and autoscaling loops.

\subsection{RQ3: Robustness under Non-Rigid Periodicity}
\label{sec:failure_modes}

We next evaluate fixed periodic templates under controlled NRP using synthetic datasets where the cycle-state dynamics are known by construction.

\begin{figure*}[t]
  \centering
  \includegraphics[width=\textwidth]{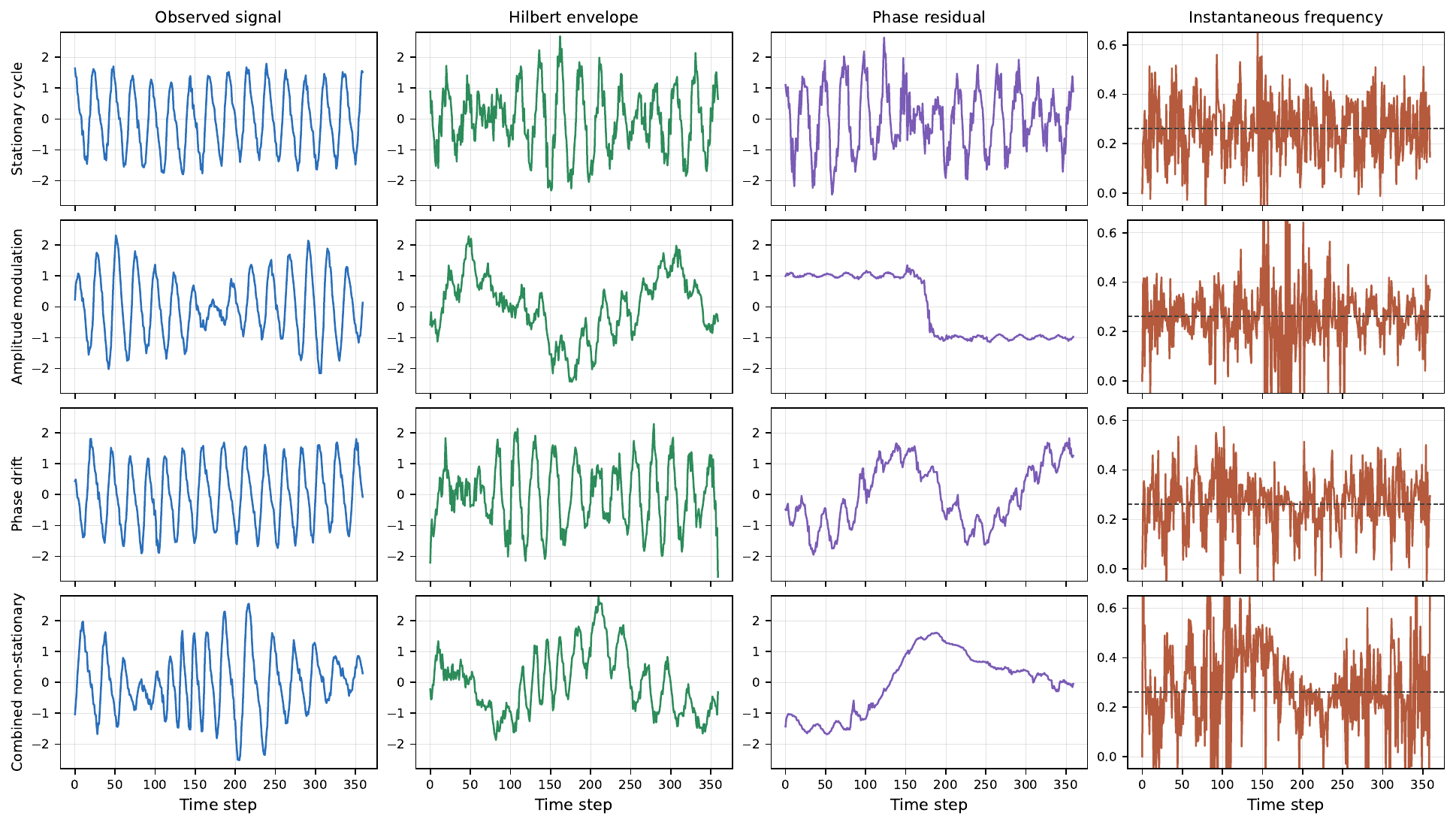}
  \caption{Controlled synthetic variants and their Hilbert-domain characterization. Each row shows a representative signal, its envelope, phase residual relative to the base period, and instantaneous frequency. Syn-A and Syn-P isolate cycle-magnitude and cycle-alignment variation, while Syn-C combines magnitude, alignment, and cycle-duration variation.}
  \label{fig:aos_synthetic_states}
\end{figure*}

\begin{table}[t]
  \centering
  \footnotesize
  \caption{Synthetic dataset specifications. All variants share the same generation framework but differ in which oscillatory-state components vary over time.}
  \label{tab:synthetic_datasets}
  \begin{adjustbox}{max width=\linewidth}
    \begin{tabular}{llccl}
      \toprule
      Alias & Variant        & Length & Channels & Cycle-state variation                              \\
      \midrule
      Syn-S & Stationary     & 12000  & 16       & None (constant $A$, $\phi$, $P$)                   \\
      Syn-A & Magnitude      & 12000  & 16       & Envelope $A_c(t)$ varies smoothly                  \\
      Syn-P & Alignment      & 12000  & 16       & Phase residual $\delta_c(t)$ drifts                \\
      Syn-C & Combined       & 12000  & 16       & $A_c(t)$, $\delta_c(t)$, and $P_c(t)$ vary jointly \\
      \bottomrule
    \end{tabular}
  \end{adjustbox}
\end{table}

\subsubsection{Setup}
Table~\ref{tab:synthetic_datasets} summarizes the four controlled datasets (base period $P_0{=}24$, $16$ channels, length $12{,}000$). Syn-S is stationary; Syn-A and Syn-P isolate cycle-magnitude and cycle-alignment variation, respectively; Syn-C evaluates the combined setting where magnitude, alignment, and cycle duration vary together. We compare \Method with CycleNet and TQNet under period settings $P\in\{12,24,36,48\}$. The detailed generation procedure is described below.

For channel $c$ and time index $t$, we define a base period $P_0=24$ and construct a time-varying amplitude $A_c(t)$, phase residual $\delta_c(t)$, and local period $P_c(t)$. The instantaneous angular frequency is
\begin{equation}
  \omega_c(t)=\frac{2\pi}{P_c(t)},
\end{equation}
and the phase is accumulated as
\begin{equation}
  \phi_c(t)=\phi_{c,0}+\sum_{\tau=1}^{t}\omega_c(\tau)+\delta_c(t),
\end{equation}
where $\phi_{c,0}$ is a channel-specific random phase. The observed value is
\begin{equation}
  \begin{aligned}
    x_c(t)
    =
    s_c\!\big[
     & A_c(t)\sin\phi_c(t)                    \\
     & + \rho_c A_c(t)\sin(2\phi_c(t)+\psi_c)
      \big]
    + r_c(t) + \epsilon_c(t),
  \end{aligned}
\end{equation}
where $s_c$ is a channel scale, $\rho_c$ controls a second harmonic, $r_c(t)$ is a weak trend and shared low-frequency component, and $\epsilon_c(t)$ is Gaussian noise.

The synthetic variants differ only in which oscillatory-state components are allowed to vary:
\begin{itemize}
  \item \textbf{Stationary cycle}: $A_c(t)=1$, $\delta_c(t)=0$, and $P_c(t)=P_0$.
  \item \textbf{Cycle-magnitude variation}: $A_c(t)$ varies smoothly while phase and period remain stable.
  \item \textbf{Cycle-alignment variation}: $\delta_c(t)$ varies smoothly and includes a small random-walk component.
  \item \textbf{Combined NRP}: amplitude, phase residual, and local period all vary simultaneously.
\end{itemize}
Thus, cycle-duration variation is evaluated within the combined setting, where it interacts with magnitude and alignment changes as it commonly does in real data. We do not report a separate duration-only variant; the synthetic analysis is intended to test whether the proposed state descriptors remain useful as non-rigid cycle dynamics become progressively less compatible with fixed template retrieval.

The experiments in Table~\ref{tab:aos_synthetic} use sequences of length $12000$ with $16$ channels. We train \Method and TQNet using the same look-back length $L=96$ and horizon $H=96$. TQNet is evaluated with period choices $\{12,24,36,48\}$, while \Method does not use a manually specified period.

\begin{table*}[t]
  \centering
  \caption{Controlled synthetic results (look-back $96$, horizons $96$ and $192$). Lower is better. CycleNet and TQNet are evaluated with period choices $P\in\{12,24,36,48\}$, while \Method uses no manually specified period.}
  \label{tab:aos_synthetic}
  \footnotesize
  \begin{adjustbox}{max width=\linewidth}
    \begin{tabular}{ll cc || cc cc cc cc || cc cc cc cc cc cc}
      \toprule
                                             &     &
      \multicolumn{2}{c}{\Method}            &
      \multicolumn{2}{c}{CycleNet/12} &
      \multicolumn{2}{c}{CycleNet/24} &
      \multicolumn{2}{c}{CycleNet/36} &
      \multicolumn{2}{c}{CycleNet/48} &
      \multicolumn{2}{c}{TQNet/12}    &
      \multicolumn{2}{c}{TQNet/24}    &
      \multicolumn{2}{c}{TQNet/36}    &
      \multicolumn{2}{c}{TQNet/48}                                                                                                                                                                                                                                                                                                                                                    \\
      \cmidrule(lr){3-4}\cmidrule(lr){5-6}\cmidrule(lr){7-8}\cmidrule(lr){9-10}\cmidrule(lr){11-12}\cmidrule(lr){13-14}\cmidrule(lr){15-16}\cmidrule(lr){17-18}\cmidrule(lr){19-20}
      Variant                                & $H$ & MSE             & MAE                        & MSE                     & MAE          & MSE          & MAE          & MSE          & MAE          & MSE   & MAE   & MSE                    & MAE                    & MSE                    & MAE                    & MSE   & MAE   & MSE                    & MAE                    \\
      \midrule
      \multirow{2}{*}{Syn-S}
                                             & 96  & \rank{1}{0.013} & \rank{1}{0.089} & 0.021 & 0.115 & \best{0.021}        & \best{0.115}        & 0.022        & 0.116        & 0.021 & 0.115 & \rank{2}{\best{0.014}} & \rank{2}{\best{0.095}} & 0.015                  & 0.098                  & 0.017 & 0.102 & 0.015                  & 0.098                  \\
                                             & 192 & \rank{1}{0.012} & \rank{1}{0.089} & 0.026        & 0.127        & \best{0.025}        & \best{0.125} & 0.025 & 0.125 & 0.026 & 0.127 & 0.014                  & 0.094                  & \rank{2}{\best{0.014}} & 0.094                  & 0.014 & 0.094 & \best{0.014} & \rank{2}{\best{0.093}} \\
      \midrule
      \multirow{2}{*}{Syn-A}
                                             & 96  & \rank{1}{0.023} & \rank{1}{0.115} & 0.055        & 0.178        & \best{0.055} & \best{0.177} & 0.057        & 0.180        & 0.056 & 0.179 & 0.028                  & 0.129                  & \rank{2}{\best{0.028}} & \rank{2}{\best{0.127}} & 0.029 & 0.129 & 0.028                  & 0.128                  \\
                                             & 192 & \rank{1}{0.024} & \rank{1}{0.116} & 0.061        & 0.191        & \best{0.056} & \best{0.184} & 0.060        & 0.189        & 0.060 & 0.190 & \rank{2}{\best{0.025}} & \rank{2}{\best{0.122}} & 0.026                  & 0.125                  & 0.025 & 0.122 & 0.028                  & 0.130                  \\
      \midrule
      \multirow{2}{*}{Syn-P}
                                             & 96  & \rank{1}{0.017} & \rank{1}{0.104} & 0.032 & 0.142 & \best{0.032}        & \best{0.142}        & 0.033        & 0.145        & 0.032 & 0.142 & 0.021                  & 0.114                  & \rank{2}{\best{0.019}} & \rank{2}{\best{0.109}} & 0.022 & 0.116 & 0.019                  & 0.110                  \\
                                             & 192 & \rank{1}{0.016} & \rank{1}{0.100} & \best{0.036} & \best{0.154} & 0.038        & 0.156        & 0.036        & 0.154        & 0.038 & 0.156 & \rank{2}{\best{0.018}} & \rank{2}{\best{0.106}} & 0.020                  & 0.111                  & 0.018 & 0.106 & 0.019                  & 0.111                  \\
      \midrule
      \multirow{2}{*}{Syn-C}
                                             & 96  & \rank{1}{0.661} & \rank{1}{0.617} & 0.737        & 0.663        & 0.736        & 0.663        & \best{0.735} & \best{0.662} & 0.738 & 0.664 & 0.671                  & 0.621                  & 0.673                  & 0.621                  & 0.668 & 0.619 & \rank{2}{\best{0.666}} & \rank{2}{\best{0.618}} \\
                                             & 192 & \rank{1}{0.746} & \rank{1}{0.669} & \best{0.817} & \best{0.709} & 0.821        & 0.711        & 0.818        & 0.710        & 0.823 & 0.711 & 0.774                  & 0.682                  & \rank{2}{\best{0.767}} & \rank{2}{\best{0.678}} & 0.771 & 0.681 & 0.777                  & 0.682                  \\
      \bottomrule
    \end{tabular}
  \end{adjustbox}
\end{table*}

\subsubsection{Results}
Table~\ref{tab:aos_synthetic} reveals a clear pattern. On the stationary baseline (Syn-S), \Method achieves the best MSE, showing that oscillatory-state alignment does not degrade performance when fixed templates are sufficient. Across increasingly complex non-rigid variants---from cycle-magnitude variation (Syn-A), to cycle-alignment variation (Syn-P), to the combined setting with magnitude, alignment, and duration changes (Syn-C)---\Method's advantage over both CycleNet and TQNet grows consistently. Notably, CycleNet's performance is largely insensitive to the period setting, indicating that its failure stems from the template assumption itself rather than period misspecification alone. TQNet's attention-based template is more expressive, yet still degrades under cycle-alignment variation (MSE changes from $0.019$ at $P{=}24$ to $0.022$ at $P{=}36$), whereas \Method requires no period specification.

\begin{figure*}[t]
  \centering
  \includegraphics[width=\textwidth]{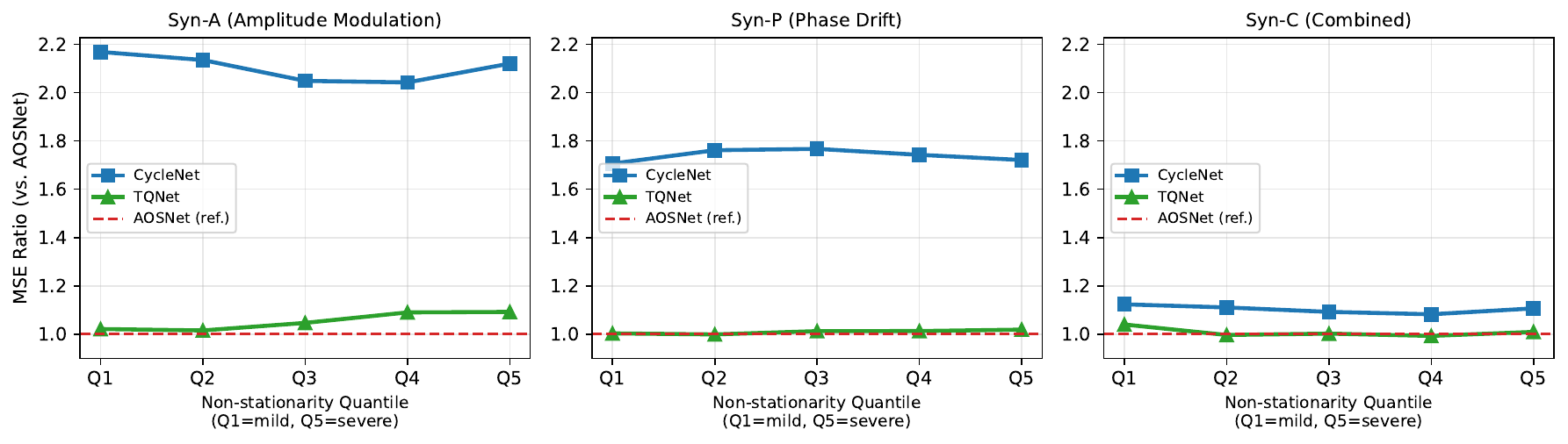}
  
  \caption{Stratified evaluation by cycle-state variation severity. Test samples are binned into five quantiles (Q1=mildest, Q5=most severe). The $y$-axis shows MSE ratio relative to \Method (dashed line at $1.0$), summarizing how relative error changes with cycle-state variation.}
  \label{fig:stratified_ns}
\end{figure*}

\subsubsection{Stratified Analysis}
To further quantify this trend, we partition the test set into five quantile bins by per-sample cycle-state variation score. Fig.~\ref{fig:stratified_ns} shows that CycleNet's error remains ${\sim}2{\times}$ that of \Method regardless of severity on Syn-A, while TQNet's ratio increases monotonically from Q1 to Q5. This supports the claim that adaptive oscillatory-state alignment becomes increasingly advantageous as the local oscillatory state deviates further from a fixed template.

\begin{figure*}[!t]
  \centering
  \includegraphics[width=\textwidth]{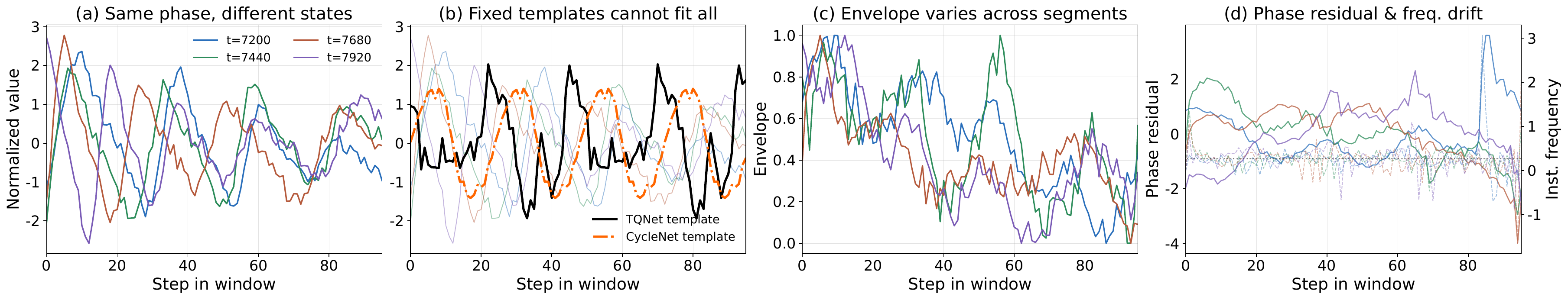}
  \caption{Phase-aligned windows and local cycle states on the combined non-rigid dataset (Syn-C). (a)~Windows starting from the same nominal cycle phase exhibit different local shapes. (b)~TQNet and CycleNet assign fixed templates to phase-aligned windows. (c)--(d)~Hilbert descriptors show differences in envelope, phase residual, and instantaneous frequency across these windows.}
  \label{fig:template_vs_aos_prior}
\end{figure*}

\begin{figure*}[!t]
  \centering
  \includegraphics[width=0.95\textwidth]{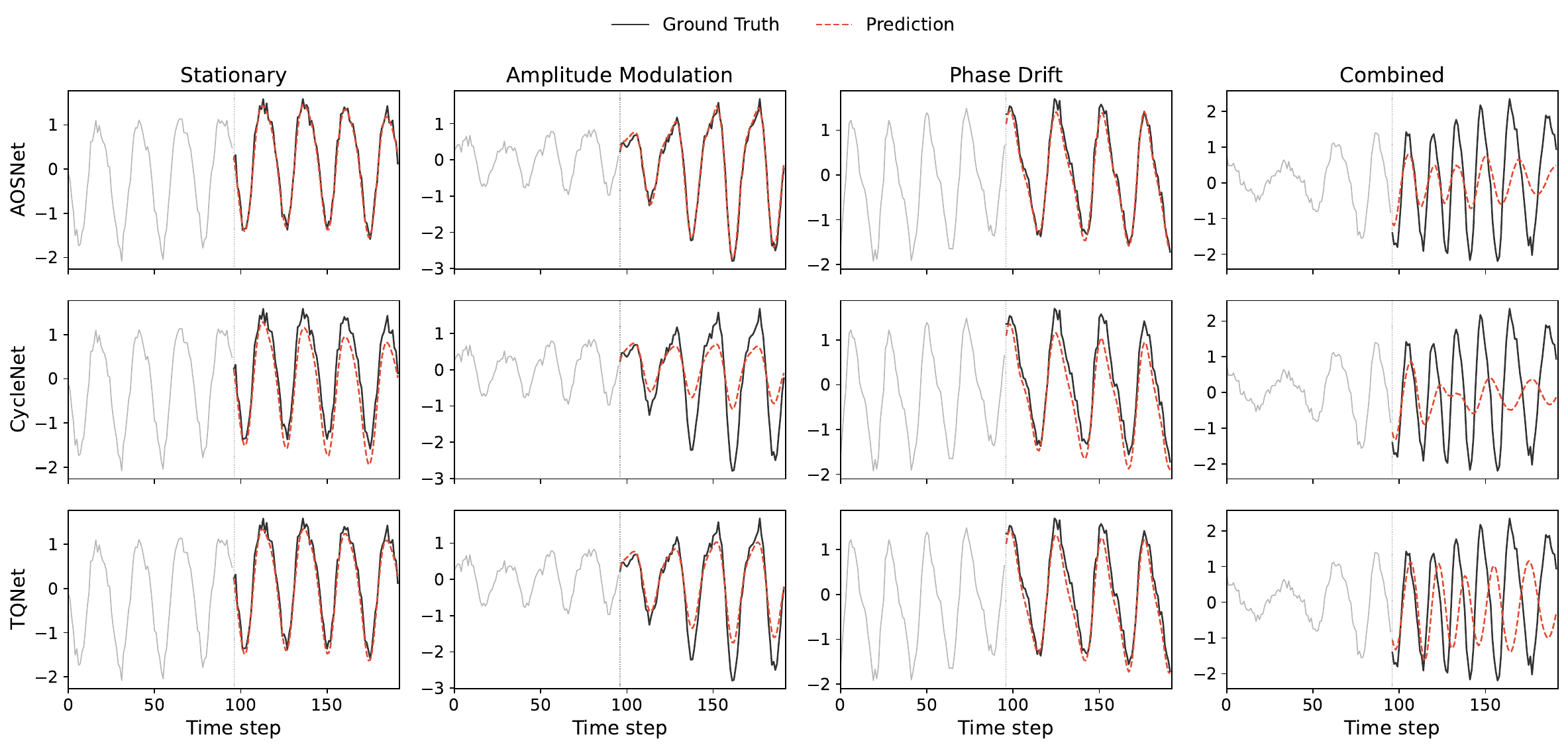}
  \caption{Prediction visualization on four synthetic datasets (look-back $96$, horizon $96$). The panels compare model forecasts under stationary cycles, envelope changes (Syn-A), phase shifts (Syn-P), and their combination (Syn-C).}
  \label{fig:synthetic_prediction}
\end{figure*}

\subsubsection{Case Study: Same Nominal Phase, Different Cycle States}
\label{sec:case_study}

Fig.~\ref{fig:template_vs_aos_prior} illustrates the difference between fixed-template periodicity and oscillatory-state alignment. Multiple observed windows in the Syn-C dataset start from the same nominal cycle phase, yet their local cycle states---as revealed by envelope, phase residual, and instantaneous frequency---differ substantially. A period-indexed method (CycleNet, TQNet) must assign identical templates to all phase-aligned windows, because the template is retrieved solely by the period index. In contrast, \Method reads the local oscillatory state of each window and adapts its alignment accordingly. This illustrates the core insight: \emph{same nominal phase does not imply the same cycle state}, which is precisely why fixed periodic indexing fails under NRP.

Fig.~\ref{fig:synthetic_prediction} further visualizes representative forecasts. On Syn-A and Syn-P, the fixed-template baselines produce periodic predictions that diverge when the local state deviates from the training-set average, whereas \Method adapts its forecast to the instantaneous oscillatory state.

\subsection{RQ4: Ablation and Sensitivity}
\label{sec:ablation}

\subsubsection{Component Ablation}

\begin{table}[t]
  \centering
  \small
  \setlength{\tabcolsep}{3pt}
  \caption{Ablation study (look-back $96$, horizon $96$). $\Delta$\% denotes relative MSE degradation from the full model.}
  \label{tab:ablation}
  \begin{adjustbox}{max width=\linewidth}
    \begin{tabular}{l ccc ccc ccc ccc}
      \toprule
                     & \multicolumn{3}{c}{ETTh1} & \multicolumn{3}{c}{ETTm1} & \multicolumn{3}{c}{Electricity} & \multicolumn{3}{c}{Weather}                                                                                                                             \\
      \cmidrule(lr){2-4} \cmidrule(lr){5-7} \cmidrule(lr){8-10} \cmidrule(lr){11-13}
      Variant        & MSE                       & MAE                       & $\Delta$\%                      & MSE                         & MAE            & $\Delta$\% & MSE            & MAE            & $\Delta$\% & MSE            & MAE            & $\Delta$\% \\
      \midrule
      \Method (full) & \textbf{0.368}            & \textbf{0.392}            & --                              & \textbf{0.307}              & \textbf{0.345} & --         & \textbf{0.132} & \textbf{0.225} & --         & \textbf{0.148} & \textbf{0.189} & --         \\
      \midrule
      w/o AOS        & 0.380                     & 0.398                     & +3.3                            & 0.319                       & 0.353          & +3.9       & 0.145          & 0.234          & +9.8       & 0.169          & 0.206          & +14.2      \\
      w/o Prior      & 0.375                     & 0.393                     & +1.9                            & 0.321                       & 0.355          & +4.6       & 0.145          & 0.234          & +9.8       & 0.171          & 0.208          & +15.5      \\
      w/o MHA        & 0.365                     & 0.390                     & $-$0.8                          & 0.315                       & 0.350          & +2.6       & 0.150          & 0.238          & +13.6      & 0.162          & 0.202          & +9.5       \\
      w/o Fusion     & 0.380                     & 0.389                     & +3.3                            & 0.327                       & 0.358          & +6.5       & 0.180          & 0.265          & +36.4      & 0.165          & 0.208          & +11.5      \\
      \midrule
      w/o Amp        & 0.370                     & 0.393                     & +0.5                            & 0.316                       & 0.348          & +2.9       & 0.133          & 0.227          & +1.5       & 0.153          & 0.195          & +3.4       \\
      w/o Phase      & 0.371                     & 0.393                     & +0.8                            & 0.313                       & 0.348          & +2.0       & 0.135          & 0.228          & +2.3       & 0.149          & 0.190          & +0.5       \\
      w/o IF         & 0.369                     & 0.390                     & +0.3                            & 0.315                       & 0.350          & +2.6       & 0.134          & 0.228          & +1.5       & 0.149          & 0.192          & +0.5       \\
      \bottomrule
    \end{tabular}
  \end{adjustbox}
\end{table}

Table~\ref{tab:ablation} evaluates the contribution of each component. The AOS module is consistently beneficial ($+3.3\%$ to $+14.2\%$), with the largest gains on datasets with complex oscillatory dynamics (Weather $+14.2\%$, Electricity $+9.8\%$). Removing the global oscillatory prior shows a similar pattern ($+15.5\%$ on Weather), indicating that the learned reference provides useful oscillatory-state information. Dual-path fusion is critical on high-dimensional data (Electricity $+36.4\%$), while cross-variate MHA is indispensable on multi-channel datasets ($+13.6\%$ on Electricity) but neutral on the $7$-channel ETTh1. Among the three analytic-signal descriptors, all contribute complementarily, with log-amplitude most impactful on Weather ($+3.4\%$).

\subsubsection{Sensitivity}
\label{sec:sensitivity}

\begin{figure}[!t]
  \centering
  \includegraphics[width=\linewidth]{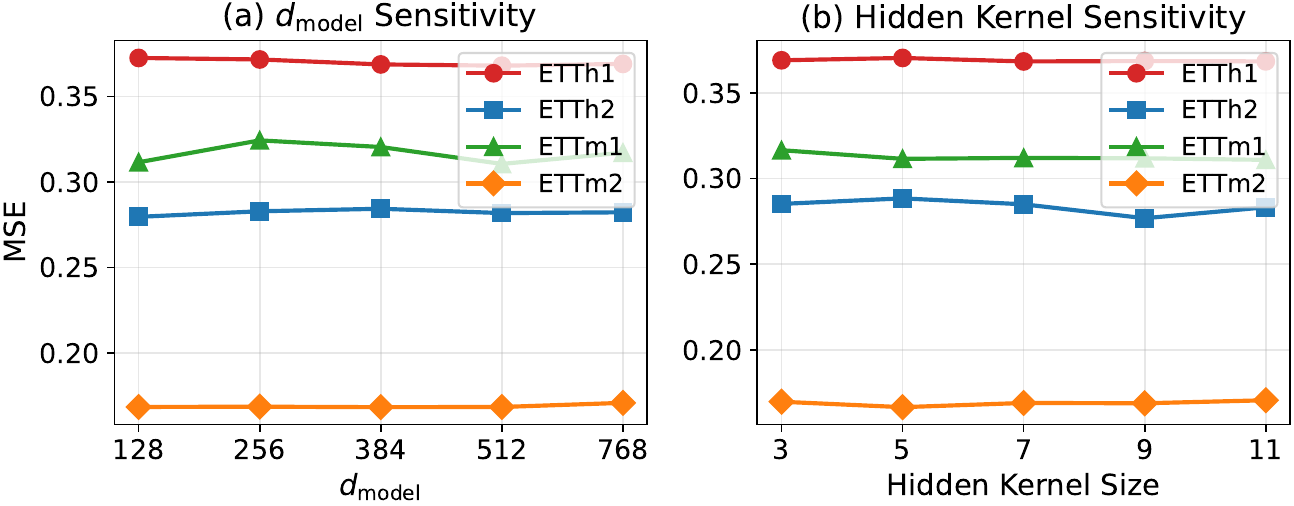}
  \vspace{-15pt}
  \caption{Parameter sensitivity on four ETT datasets (horizon $96$). (a)~Varying $d_{\mathrm{model}}\in\{128,256,384,512,768\}$ with kernel fixed at $5$. (b)~Varying kernel size $\in\{3,5,7,9,11\}$ with $d_{\mathrm{model}}{=}512$. Performance remains within ${\pm}5\%$ across all settings.}
  \label{fig:sensitivity}
  \vspace{6pt}
  \includegraphics[width=\linewidth]{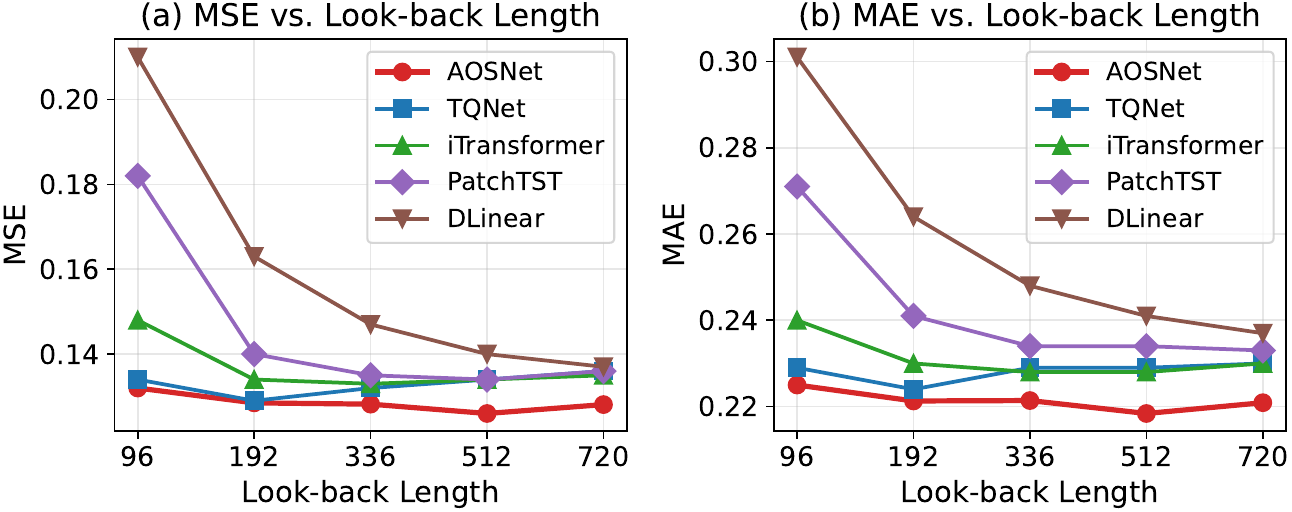}
  \vspace{-15pt}
  \caption{Look-back length sensitivity on Electricity (horizon $96$). \Method achieves the lowest MSE at every tested length and already performs strongly at $L{=}96$, whereas PatchTST and DLinear require substantially longer windows to become competitive.}
  \label{fig:lookback_sensitivity}
\end{figure}

Fig.~\ref{fig:sensitivity} shows that \Method is robust to both model dimension and kernel size, with performance remaining nearly flat across the full tested range. Even reducing $d_{\mathrm{model}}$ to $128$ ($4{\times}$ fewer parameters) incurs at most ${\sim}1\%$ MSE increase, making the model suitable for resource-constrained deployment. Fig.~\ref{fig:lookback_sensitivity} further shows that \Method achieves the best MSE at every tested look-back length ($96$--$720$) on Electricity. All methods improve with longer windows as expected; \Method already performs strongly at the shortest setting ($L{=}96$), while PatchTST and DLinear require $L{\geq}336$ to approach comparable accuracy.

\subsection{RQ5: Mechanism Analysis of Oscillatory-State Alignment}
\label{sec:alignment_analysis}

We now examine the internal mechanism of \Method to probe whether the model operates through oscillatory-state alignment rather than simply adding signal-processing features.

\begin{figure}[t]
  \centering
  \includegraphics[width=0.94\columnwidth]{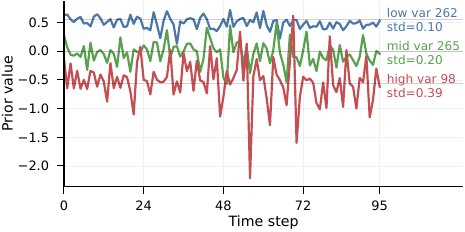}
  \caption{Learned global oscillatory prior on Electricity. Variables are selected by prior standard deviation (low, median, high oscillation strength). The traces show uneven amplitudes and nonuniform local shapes.}
  \label{fig:prior_traces}
\end{figure}

\begin{figure}[t]
  \centering
  \includegraphics[width=0.94\columnwidth]{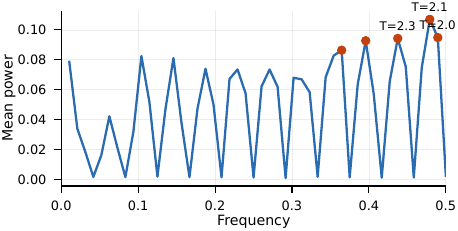}
  \caption{Average spectrum of the learned prior. Multiple spectral peaks indicate that the prior captures a mixture of oscillatory scales rather than committing to a single predefined period.}
  \label{fig:prior_spectrum}
\end{figure}

\subsubsection{The Prior Acts as a Flexible Reference}
Fig.~\ref{fig:prior_traces} visualizes representative prior traces from the trained Electricity model. The curves exhibit oscillatory behavior, as expected because the prior is optimized to provide amplitude, phase, and frequency references, but their amplitudes are uneven and local shapes vary. Fig.~\ref{fig:prior_spectrum} further shows multiple spectral peaks, indicating that the prior stores a mixture of oscillatory scales rather than a single repeated cycle. These observations support our interpretation that the global prior serves as a flexible reference. Importantly, we do not claim that a learnable length-$L$ waveform alone resolves the limitation of fixed templates; the distinction is that its influence is locally modulated by Hilbert-domain state descriptors rather than retrieved by a period index.

\begin{figure}[t]
  \centering
  \includegraphics[width=0.94\columnwidth]{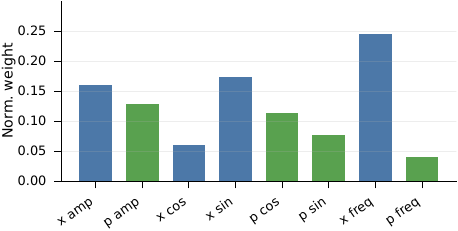}
  \caption{Effective descriptor weights in the adaptive gate. The gate assigns large weights to observed instantaneous frequency, phase, and amplitude, while using prior descriptors as the reference state for comparison.}
  \label{fig:gate_descriptor_weights}
\end{figure}
\balance

\subsubsection{The Gate Operates in Oscillatory-State Space}
Fig.~\ref{fig:gate_descriptor_weights} shows that the adaptive gate assigns large effective weights to observed instantaneous frequency, phase, and amplitude descriptors, while prior descriptors provide the reference state. This pattern is consistent with a Hilbert-domain state comparison rather than unconditional copying of the prior. The learned fusion coefficient is $\lambda{=}0.100$, meaning that about $90\%$ of the fused prediction is contributed by the base projection of the aligned sequence, suggesting that the aligned temporal signal itself remains the main forecasting substrate.

\begin{figure}[t]
  \centering
  \includegraphics[width=0.94\columnwidth]{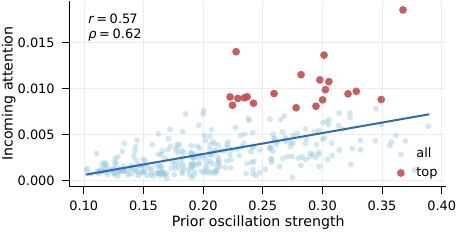}
  \caption{Relation between prior oscillation strength and incoming channel attention. The plot compares each variable's prior oscillation strength with its incoming cross-variate attention.}
  \label{fig:attention_prior_relation}
\end{figure}

\subsubsection{Oscillatory States Structure Cross-Variate Attention}
Fig.~\ref{fig:attention_prior_relation} shows that the cross-variate MHA tends to assign larger incoming attention to channels with stronger learned oscillatory priors. Combined with the ASR analysis (Table~\ref{tab:asr}), this indicates that per-channel oscillatory-state alignment produces structurally refined representations that help the downstream attention discover meaningful cross-variate dependencies.

\subsubsection{Cross-Variate Dependency Analysis}

To quantify how oscillatory-state alignment structures cross-variate representations, we compute the Attention Structure Ratio (ASR):
\begin{equation}
  \text{ASR} = \frac{\bar{A}_{\text{intra}}}{\bar{A}_{\text{inter}}},
  \label{eq:asr}
\end{equation}
where $\bar{A}_{\text{intra}}$ and $\bar{A}_{\text{inter}}$ denote average attention weights within and between K-Means clusters of channel embeddings.

\begin{table}[t]
  \centering
  \footnotesize
  \caption{Attention Structure Ratio (ASR) with and without AOS. ASR $>1$ indicates structured cross-variate attention; ASR $\leq 1$ indicates uniform attention.}
  \label{tab:asr}
  \begin{adjustbox}{max width=\linewidth}
    \begin{tabular}{lcccc}
      \toprule
                     & \multicolumn{2}{c}{Electricity (321 ch.)} & \multicolumn{2}{c}{Weather (21 ch.)}                                                                               \\
      \cmidrule(lr){2-3} \cmidrule(lr){4-5}
      Variant        & ASR                                       & $\bar{A}_{\text{intra}}/\bar{A}_{\text{inter}}$ & ASR            & $\bar{A}_{\text{intra}}/\bar{A}_{\text{inter}}$ \\
      \midrule
      \Method (full) & \textbf{1.208}                            & 0.00342\,/\,0.00283                             & \textbf{1.213} & 0.0532\,/\,0.0439                               \\
      w/o AOS        & 0.716                                     & 0.00257\,/\,0.00359                             & 0.971          & 0.0472\,/\,0.0486                               \\
      \bottomrule
    \end{tabular}
  \end{adjustbox}
\end{table}

As shown in Table~\ref{tab:asr}, the full model achieves ASR $>1.2$ on both datasets, while removing AOS causes ASR to drop below $1.0$ on Electricity. This indicates that oscillatory-state alignment is critical for structuring the input so that the MHA can discover cross-variate dependencies.

\section{Discussion}
\label{sec:discussion}

These results suggest that oscillatory-state alignment is most useful when recurrence remains present but its local state is no longer well represented by a fixed phase-indexed template. At the same time, several limitations remain. (1) The current study focuses on point forecasting with a compact backbone, so the benefit of adaptive oscillatory states has not yet been fully explored in larger Transformer-style architectures or probabilistic forecasting settings. (2) The Hilbert descriptors are computed along fixed look-back windows, which may be less suitable when useful oscillatory structure appears at multiple resolutions or across irregularly sampled observations. (3) The global temporal prior is shared across samples; while this provides a simple dataset-level reference, future work could make the prior conditional on covariates, regimes, or cluster-level temporal states. (4) The current experiments emphasize standard benchmarks and controlled non-rigid periodic variants. A broader evaluation on domain-specific datasets with abrupt regime changes, missing values, and exogenous events would further clarify when adaptive oscillatory-state alignment is most beneficial.

\section{Conclusion}
\label{sec:conclusion}

We presented \Method, a Hilbert-guided forecasting model that reformulates explicit periodic modeling from fixed template matching to adaptive oscillatory-state alignment. Instead of requiring a predefined period or a globally stable repeated pattern, \Method extracts envelope, phase, and instantaneous-frequency descriptors from both the observed sequence and a learnable global temporal prior. A lightweight adaptive gate then uses these local descriptors to decide where the observation should be preserved and where it should be softly corrected toward the learned reference. Experiments on public benchmarks, cloud workload traces, and controlled synthetic variants show that this design is especially useful under NRP, where cycle magnitude, alignment, or cycle duration changes over time, and model analysis suggests that the learned prior acts as an oscillatory reference rather than a hard periodic template.

\clearpage
\bibliographystyle{IEEEtran}
\bibliography{refs}

@article{Wen-2022-Transformerstimeseries,
  author  = {Wen, Qingsong and Zhou, Tian and Zhang, Chaoli and Chen, Weiqi and Ma, Ziqing and Yan, Junchi and Sun, Liang},
  journal = {arXiv preprint arXiv:2202.07125},
  title   = {{Transformers in Time Series: A Survey}},
  year    = {2022}
}

@inproceedings{Qiu-2024-TFB,
  address   = {Guangzhou, China, Aug. 26-30},
  author    = {Xiangfei Qiu and Jilin Hu and Lekui Zhou and Xingjian Wu and Junyang Du and Buang Zhang and Chenjuan Guo and Aoying Zhou and Christian S. Jensen and Zhenli Sheng and Bin Yang},
  booktitle = {International Conference on Very Large Data Bases (VLDB)},
  title     = {{TFB: Towards Comprehensive and Fair Benchmarking of Time Series Forecasting Methods}},
  year      = {2024}
}

@article{Shao-2025-ExploringProgressMultivariate,
  author    = {Zezhi Shao and Fei Wang and Yongjun Xu and Wei Wei and Chengqing Yu and Zhao Zhang and Di Yao and Tao Sun and Guangyin Jin and Xin Cao and Gao Cong and Christian S. Jensen and Xueqi Cheng},
  doi       = {10.1109/TKDE.2024.3484454},
  issn      = {2326-3865},
  issue     = {1},
  journal   = {IEEE Transactions on Knowledge and Data Engineering},
  month     = jan,
  pages     = {291--305},
  publisher = {IEEE},
  title     = {{Exploring Progress in Multivariate Time Series Forecasting: Comprehensive Benchmarking and Heterogeneity Analysis}},
  volume    = {37},
  year      = {2025}
}

@article{Guo-2026-SurveyDeepLearning_TSF,
  author  = {Guo, Qi and Zhao, Baolin and Song, Mingchen and Zhong, Guoqiang},
  doi     = {10.1109/TKDE.2026.3671378},
  journal = {IEEE Transactions on Knowledge and Data Engineering},
  pages   = {1-20},
  title   = {{A Survey of Deep Learning for Time Series Forecasting: Taxonomy, Analysis and Future Directions}},
  year    = {2026}
}

@inproceedings{Zhou-2021-Informerefficienttransformer,
  address   = {Virtual Event, Feb. 2-9},
  author    = {Zhou, Haoyi and Zhang, Shanghang and Peng, Jieqi and Zhang, Shuai and Li, Jianxin and Xiong, Hui and Zhang, Wancai},
  booktitle = {Proceedings of The AAAI Conference on Artificial Intelligence},
  title     = {{Informer: Beyond Efficient Transformer for Long Sequence Time-Series Forecasting}},
  year      = {2021}
}

@inproceedings{Zeng-2023-DLinear,
  address   = {Washington, DC, USA, Feb. 7-14},
  author    = {Zeng, Ailing and Chen, Muxi and Zhang, Lei and Xu, Qiang},
  booktitle = {Proceedings of The AAAI Conference on Artificial Intelligence},
  title     = {{Are Transformers Effective for Time Series Forecasting?}},
  year      = {2023}
}

@inproceedings{Zhang-2025-MultiPeriodLearning,
  address       = {Toronto, ON, Canada, Aug. 3-7},
  archiveprefix = {arXiv},
  author        = {Zhang, Xu and Huang, Zhengang and Wu, Yunzhi and Lu, Xun and Qi, Erpeng and Chen, Yunkai and Xue, Zhongya and Wang, Qitong and Wang, Peng and Wang, Wei},
  booktitle     = {Proceedings of The 31st ACM SIGKDD Conference on Knowledge Discovery and Data Mining},
  doi           = {10.1145/3690624.3709422},
  eprint        = {2511.08622},
  month         = jul,
  pages         = {2848--2859},
  primaryclass  = {q-fin},
  title         = {{Multi-Period Learning for Financial Time Series Forecasting}},
  url           = {http://arxiv.org/abs/2511.08622},
  urldate       = {2025-11-13},
  year          = {2025}
}

@article{Chen-2023-FuXi,
  author       = {Chen, Lei and Zhong, Xiaohui and Zhang, Feng and Cheng, Yuan and Xu, Yinghui and Qi, Yuan and Li, Hao},
  doi          = {10.1038/s41612-023-00512-1},
  issn         = {2397-3722},
  journal      = {npj Climate and Atmospheric Science},
  month        = nov,
  number       = {1},
  pages        = {190},
  shortjournal = {npj Clim Atmos Sci},
  shorttitle   = {{FuXi}},
  title        = {{FuXi: A Cascade Machine Learning Forecasting System for 15-Day Global Weather Forecast}},
  urldate      = {2025-10-20},
  volume       = {6},
  year         = {2023}
}

@inproceedings{Wu-2021-Autoformer,
  address   = {Virtual, Dec. 6-14},
  author    = {Wu, Haixu and Xu, Jiehui and Wang, Jianmin and Long, Mingsheng},
  booktitle = {Annual Conference on Neural Information Processing Systems (NeurIPS)},
  title     = {{Autoformer: Decomposition Transformers with Auto-Correlation for Long-Term Series Forecasting}},
  year      = {2021}
}

@inproceedings{Nie-2023-PacthTST,
  address   = {Kigali, Rwanda, May 1-5},
  author    = {Nie, Yuqi and H. Nguyen, Nam and Sinthong, Phanwadee and Kalagnanam, Jayant},
  booktitle = {International Conference on Learning Representations (ICLR)},
  title     = {{A Time Series Is Worth 64 Words: Long-Term Forecasting with Transformers}},
  year      = {2023}
}

@inproceedings{Liu-2024-iTransformer,
  address   = {Vienna, Austria, May 7-11},
  author    = {Yong Liu and Tengge Hu and Haoran Zhang and Haixu Wu and Shiyu Wang and Lintao Ma and Mingsheng Long},
  booktitle = {International Conference on Learning Representations (ICLR)},
  title     = {{{iTransformer}: Inverted Transformers Are Effective for Time Series Forecasting}},
  year      = {2024}
}

@inproceedings{Ye-2024-FAN,
  address   = {Vancouver, BC, Canada, Dec. 10 - 15},
  author    = {Weiwei Ye and Songgaojun Deng and Qiaosha Zou and Ning Gui},
  booktitle = {Annual Conference on Neural Information Processing Systems (NeurIPS)},
  title     = {{Frequency Adaptive Normalization for Non-Stationary Time Series Forecasting}},
  year      = {2024}
}

@inproceedings{Liu-2025-TimeBridge,
  address   = {Vancouver, Canada, Jul. 13-19},
  author    = {Liu, Peiyuan and Wu, Beiliang and Hu, Yifan and Li, Naiqi and Dai, Tao and Bao, Jigang and Xia, Shu-Tao},
  booktitle = {International Conference on Machine Learning (ICML)},
  title     = {{TimeBridge: Non-Stationarity Matters for Long-Term Time Series Forecasting}},
  year      = {2025}
}

@article{Song-2026-DSTND,
  author  = {Song, Zhangyao and Zhang, Xiang and Zhuang, Li and Guo, Tao and Zhao, Xiaoyu and Xu, Yinfei and Jin, Shi},
  doi     = {10.1109/TCCN.2026.3685404},
  journal = {IEEE Transactions on Cognitive Communications and Networking},
  number  = {},
  pages   = {7647-7661},
  title   = {Diffusion-Based Spatio-Temporal Channel Prediction via Non-Stationarity Decoupling},
  volume  = {12},
  year    = {2026}
}

@inproceedings{Lin-2024-CycleNet,
  address   = {Vancouver, Canada, Dec. 9-15},
  author    = {Lin, Shengsheng and Lin, Weiwei and Hu, Xinyi and Wu, Wentai and Mo, Ruichao and Zhong, Haocheng},
  booktitle = {Annual Conference on Neural Information Processing Systems (NeurIPS)},
  title     = {{{CycleNet}: Enhancing Time Series Forecasting through Modeling Periodic Patterns}},
  year      = {2024}
}

@inproceedings{Lin-2025-TQNet,
  address   = {BC, Canada, Jul. 13-19},
  author    = {Shengsheng Lin and Haojun Chen and Haijie Wu and Chunyun Qiu and Weiwei Lin},
  booktitle = {International Conference on Machine Learning (ICML)},
  publisher = {{PMLR} / OpenReview.net},
  series    = {Proceedings of Machine Learning Research},
  title     = {{Temporal Query Network for Efficient Multivariate Time Series Forecasting}},
  url       = {https://proceedings.mlr.press/v267/lin25e.html},
  volume    = {267},
  year      = {2025}
}

@inproceedings{Zhou-2022-FILM,
  address   = {LA, USA, Nov. 28-Dec. 9},
  author    = {Zhou, Tian and Ma, Ziqing and Wen, Qingsong and Sun, Liang and Yao, Tao and Yin, Wotao and Jin, Rong and others},
  booktitle = {Annual Conference on Neural Information Processing Systems (NeurIPS)},
  title     = {{{FILM}: Frequency Improved Legendre Memory Model for Long-Term Time Series Forecasting}},
  year      = {2022}
}

@inproceedings{Wang-2025-FreDF,
  address   = {Singapore, Apr. 24-28},
  author    = {Hao Wang and Lichen Pan and Yuan Shen and Zhichao Chen and Degui Yang and Yifei Yang and Sen Zhang and Xinggao Liu and Haoxuan Li and Dacheng Tao},
  booktitle = {International Conference on Learning Representations (ICLR)},
  title     = {{FreDF: Learning to Forecast in The Frequency Domain}},
  url       = {https://openreview.net/forum?id=4A9IdSa1ul},
  year      = {2025}
}

@inproceedings{Fei-2025-Amplifier,
  address   = {Philadelphia, PA, USA, Feb. 25 - Mar. 4},
  author    = {Jingru Fei and Kun Yi and Wei Fan and Qi Zhang and Zhendong Niu},
  booktitle = {Proceedings of The AAAI Conference on Artificial Intelligence},
  title     = {{Amplifier: Bringing Attention to Neglected Low-Energy Components in Time Series Forecasting}},
  url       = {https://doi.org/10.1609/aaai.v39i11.33267},
  year      = {2025}
}

@inproceedings{Wang-2024-TimeMixer,
  address   = {Vienna, Austria, May 7-11},
  author    = {Shiyu Wang and Haixu Wu and Xiaoming Shi and Tengge Hu and Huakun Luo and Lintao Ma and James Y. Zhang and Jun Zhou},
  booktitle = {International Conference on Learning Representations (ICLR)},
  title     = {{TimeMixer: Decomposable Multiscale Mixing for Time Series Forecasting}},
  year      = {2024}
}

@inproceedings{Yu-2024-Leddam,
  address   = {Vienna, Austria, Jul. 21-27},
  author    = {Guoqi Yu and Jing Zou and Xiaowei Hu and Angelica I Aviles-Rivero and Jing Qin and Shujun Wang},
  booktitle = {International Conference on Machine Learning (ICML)},
  title     = {{Revitalizing Multivariate Time Series Forecasting: Learnable Decomposition with {Inter-Series} Dependencies and Intra-Series Variations Modeling}},
  year      = {2024}
}

@inproceedings{Deng-2024-ParsimonyCapabilityDecomposition,
  address   = {Vancouver, Canada, Dec. 9-15},
  author    = {Jinliang Deng and Feiyang Ye and Du Yin and Xuan Song and I. Tsang and Hui Xiong},
  booktitle = {Annual Conference on Neural Information Processing Systems (NeurIPS)},
  title     = {{Parsimony or Capability? Decomposition Delivers Both in Long-Term Time Series Forecasting}},
  year      = {2024}
}

@article{Gabor-1946-TheoryCommunication,
  author  = {Gabor, Dennis},
  journal = {Journal of the Institution of Electrical Engineers},
  number  = {26},
  pages   = {429--457},
  title   = {{Theory of Communication}},
  volume  = {93},
  year    = {1946}
}

@article{Boashash-1992-EstimatingInterpreting,
  author  = {Boashash, Boualem},
  journal = {Proceedings of the IEEE},
  number  = {4},
  pages   = {520--568},
  title   = {{Estimating and Interpreting the Instantaneous Frequency of a Signal. I. Fundamentals}},
  volume  = {80},
  year    = {1992}
}

@inproceedings{Zhou-2022-FEDFormer,
  address   = {Baltimore, MD, Jul. 17-23},
  author    = {Zhou, Tian and Ma, Ziqing and Wen, Qingsong and Wang, Xue and Sun, Liang and Jin, Rong},
  booktitle = {International Conference on Machine Learning (ICML),},
  title     = {{{FEDformer}: Frequency Enhanced Decomposed Transformer for Long-Term Series Forecasting}},
  year      = {2022}
}

@inproceedings{Wang-2023-MICN,
  address   = {Kigali, Rwanda, May 1-5},
  author    = {Huiqiang Wang and Jian Peng and Feihu Huang and Jince Wang and Junhui Chen and Yifei Xiao},
  booktitle = {International Conference on Learning Representations (ICLR)},
  title     = {{{MICN}: Multi-Scale Local and Global Context Modeling for Long-Term Series Forecasting}},
  year      = {2023}
}

@inproceedings{Lin-2024-SparseTSF,
  address   = {Vienna, Austria, Jul. 21-27},
  author    = {Shengsheng Lin and Weiwei Lin and Wentai Wu and Haojun Chen and Junjie Yang},
  bibsource = {dblp computer science bibliography, https://dblp.org},
  booktitle = {International Conference on Machine Learning (ICML)},
  pages     = {30211--30226},
  publisher = {{PMLR} / OpenReview.net},
  series    = {Proceedings of Machine Learning Research},
  title     = {{SparseTSF: Modeling Long-Term Time Series Forecasting with 1K Parameters}},
  url       = {https://proceedings.mlr.press/v235/lin24n.html},
  volume    = {235},
  year      = {2024}
}

@inproceedings{Ma-2025-MoFo,
  address   = {San Diego, CA, USA, Dec. 2-7},
  author    = {Jiaming Ma and Binwu Wang and Qihe Huang and Guanjun Wang and Pengkun Wang and Zhengyang Zhou and Yang Wang},
  booktitle = {Annual Conference on Neural Information Processing Systems (NeurIPS)},
  title     = {{MoFo: Empowering Long-Term Time Series Forecasting with Periodic Pattern Modeling}},
  url       = {https://openreview.net/forum?id=sbvLts2HqR},
  year      = {2026}
}

@inproceedings{Xu-2024-FITS,
  address   = {Vienna, Austria, May 7-11},
  author    = {Xu, Zhijian and Zeng, Ailing and Xu, Qiang},
  booktitle = {International Conference on Learning Representations (ICLR)},
  title     = {{{FITS}: Modeling Time Series with 10K Parameters}},
  year      = {2024}
}

@inproceedings{Vaswani-2017-transformer,
  address   = {Long Beach, California, USA, Dec. 4-9},
  author    = {Vaswani, Ashish and Shazeer, Noam and Parmar, Niki and Uszkoreit, Jakob and Jones, Llion and Gomez, Aidan N. and Kaiser, Lukasz and Polosukhin, Illia},
  booktitle = {Annual Conference on Neural Information Processing Systems (NeurIPS)},
  title     = {{Attention {{Is All You Need}}}},
  year      = {2017}
}

@inproceedings{Song-2026-CTPNet,
  address   = {Barcelona, Spain, May 4-8},
  author    = {Song, Zhangyao and Jiang, Nanqing and He, Miaohong and Zhao, Xiaoyu and Guo, Tao},
  booktitle = {International Conference on Acoustics, Speech and Signal Processing (ICASSP)},
  doi       = {10.1109/ICASSP55912.2026.11464481},
  pages     = {4821--4825},
  publisher = {{IEEE}},
  title     = {{Channel, Trend and Periodic-Wise Representation Learning for Multivariate Long-Term Time Series Forecasting}},
  year      = {2026}
}

@article{Han-2024-capacityrobustnesstrade,
  author    = {Han, Lu and Ye, Han-Jia and Zhan, De-Chuan},
  journal   = {IEEE Transactions on Knowledge and Data Engineering},
  publisher = {IEEE},
  title     = {{The Capacity and Robustness Trade-Off: Revisiting The Channel Independent Strategy for Multivariate Time Series Forecasting}},
  year      = {2024}
}

@inproceedings{Zhang-2023-Crossformer,
  address   = {Kigali, Rwanda, May 1-5},
  author    = {Yunhao Zhang and Junchi Yan},
  booktitle = {International Conference on Learning Representations (ICLR)},
  publisher = {OpenReview.net},
  title     = {{Crossformer: Transformer Utilizing Cross-Dimension Dependency for Multivariate Time Series Forecasting}},
  url       = {https://openreview.net/forum?id=vSVLM2j9eie},
  year      = {2023}
}

@inproceedings{Huang-2024-HDMixer,
  address   = {Vancouver, Canada, Feb. 20-27},
  author    = {Huang, Qihe and Shen, Lei and Zhang, Ruixin and Cheng, Jiahuan and Ding, Shouhong and Zhou, Zhengyang and Wang, Yang},
  booktitle = {Proceedings of The AAAI Conference on Artificial Intelligence},
  title     = {{{HDMixer}: Hierarchical Dependency with Extendable Patch for Multivariate Time Series Forecasting}},
  year      = {2024}
}

@inproceedings{Ilbert-2024-SAMformer,
  address   = {Vienna, Austria, Jul. 21-27},
  author    = {Ilbert, Romain and Odonnat, Ambroise and Feofanov, Vasilii and Virmaux, Aladin and Paolo, Giuseppe and Palpanas, Themis and Redko, Ievgen},
  booktitle = {International Conference on Machine Learning (ICML)},
  publisher = {PMLR},
  title     = {{SAMformer: Unlocking The Potential of Transformers in Time Series Forecasting with Sharpness-Aware Minimization and Channel-Wise Attention}},
  url       = {https://proceedings.mlr.press/v235/ilbert24a.html},
  volume    = {235},
  year      = {2024}
}

@article{Hendrycks-2016-GELU,
  author  = {Hendrycks, Dan and Gimpel, Kevin},
  journal = {arXiv preprint arXiv:1606.08415},
  title   = {{Gaussian Error Linear Units (GELUs)}},
  year    = {2016}
}

@article{Srivastava-2014-Dropout,
  author  = {Srivastava, Nitish and Hinton, Geoffrey and Krizhevsky, Alex and Sutskever, Ilya and Salakhutdinov, Ruslan},
  journal = {Journal of Machine Learning Research},
  number  = {56},
  pages   = {1929--1958},
  title   = {{Dropout: A Simple Way to Prevent Neural Networks from Overfitting}},
  volume  = {15},
  year    = {2014}
}

@inproceedings{Paszke-2019-Pytorch,
  address   = {Vancouver, Canada, Dec. 8-14},
  author    = {Paszke, Adam and Gross, Sam and Massa, Francisco and Lerer, Adam and Bradbury, James and Chanan, Gregory and Killeen, Trevor and Lin, Zeming and Gimelshein, Natalia and Antiga, Luca and others},
  booktitle = {Annual Conference on Neural Information Processing Systems (NeurIPS)},
  title     = {{PyTorch: An Imperative Style, High-Performance Deep Learning Library}},
  year      = {2019}
}

@inproceedings{Kim-2021-RevIN,
  address   = {Virtual Event, Austria, May 3-7},
  author    = {Kim, Taesung and Kim, Jinhee and Tae, Yunwon and Park, Cheonbok and Choi, Jang-Ho and Choo, Jaegul},
  booktitle = {International Conference on Learning Representations (ICLR)},
  title     = {{Reversible Instance Normalization for Accurate Time-Series Forecasting Against Distribution Shift}},
  year      = {2021}
}

@inproceedings{Wu-2025-SRSNet,
  address   = {Vancouver, Canada, Dec. 9-14},
  author    = {Xingjian Wu and Xiangfei Qiu and Hanyin Cheng and Zhengyu Li and Jilin Hu and Chenjuan Guo and Bin Yang},
  booktitle = {Annual Conference on Neural Information Processing Systems (NeurIPS)},
  title     = {{Enhancing Time Series Forecasting through Selective Representation Spaces: A Patch Perspective}},
  year      = {2025}
}

@inproceedings{SSformer-2026-AAAI,
  address   = {Singapore, January 20-27},
  author    = {Ying Liu and
               Bo Liu and
               Sheng Huang and
               Gang Luo and
               Wenbo Hu and
               Meng Wang and
               Richang Hong},
  booktitle = {Proceedings of The AAAI Conference on Artificial Intelligence},
  pages     = {23899--23907},
  publisher = {{AAAI} Press},
  title     = {Sparse-Scale Transformer with Bidirectional Awareness for Time Series Forecasting},
  url       = {https://doi.org/10.1609/aaai.v40i28.39566},
  year      = {2026}
}

\end{document}